\documentclass[10pt,twocolumn,letterpaper]{article}

\usepackage{cvpr} 

\usepackage[accsupp]{axessibility} 

\usepackage{graphicx}
\usepackage{hyperref}
\makeatletter
\@namedef{ver@everyshi.sty}{}
\usepackage{graphicx}
\usepackage{amsmath}
\usepackage{amssymb}
\usepackage{booktabs}
\usepackage{threeparttable}
\usepackage{multirow}
\usepackage{multicol}
\usepackage[table]{xcolor}
\usepackage{amssymb} 
\usepackage{pifont}

\usepackage{bm}
\makeatother
\usepackage{tikz}
\usepackage{pgfplots}
\usepackage{soul}
\usepackage{amsfonts}
\usepackage{color}
\usepackage{xargs}
\usepackage{enumitem}
\usepackage{subcaption}

\newcommand{\datasetname}{\textsc{3MASSIV}\xspace}
\newcommand{\concept}{\texttt{concept}\xspace}

\def\cvprPaperID{10392} 
\def\confName{CVPR}
\def\confYear{2022}
\begin{document}


\title{3MASSIV: Multilingual, Multimodal and Multi-Aspect dataset of Social Media Short Videos}

\author{
Vikram Gupta\textsuperscript{\rm 1,*},
Trisha Mittal\textsuperscript{\rm 2,*}, 
Puneet Mathur\textsuperscript{\rm 2},
Vaibhav Mishra\textsuperscript{\rm 1},\\
Mayank Maheshwari\textsuperscript{\rm 1},
Aniket Bera\textsuperscript{\rm 2},
Debdoot Mukherjee\textsuperscript{\rm 1},
Dinesh Manocha\textsuperscript{\rm 2}\\
\thanks{The first two authors contributed equally to this work.}
\textsuperscript{\rm 1}ShareChat, India\\ 
\textsuperscript{\rm 2}University of Maryland, College Park, USA\\ 
\{vikramgupta, vaibhavmishra, mayankmaheshwari, debdoot\}@sharechat.co\\ 
\{trisha, puneetm, bera, dmanocha\}@umd.edu\\ 
Project URL: \url{https://sharechat.com/research/3massiv}
}

\maketitle

\begin{abstract}
We present~\datasetname, a multilingual, multimodal and multi-aspect, expertly-annotated dataset of diverse short videos extracted from short-video social media platform - Moj. \datasetname~comprises of $50k$ short videos ($20$ seconds average duration) and $100$K unlabeled videos in $11$ different languages and captures popular short video trends like pranks, fails, romance, comedy expressed via unique audio-visual formats like self-shot videos, reaction videos, lip-synching, self-sung songs, etc. \datasetname presents an opportunity for multimodal and multilingual semantic understanding on these unique videos by annotating them for concepts, affective states, media types, and audio language. We present a thorough analysis of \datasetname~and highlight the variety and unique aspects of our dataset compared to other contemporary popular datasets with strong baselines. We also show how the social media content in \datasetname~is dynamic and temporal in nature, which can be used for semantic understanding tasks and cross-lingual analysis.
\end{abstract}
\vspace{-15pt}
\section{Introduction}
\label{sec:intro}
\textit{Semantic understanding} of videos has been a well-researched problem but still continues to garner a lot of attention from the computer vision and multimedia research communities because videos encode rich information which can be understood across different dimensions using various tasks. Notable progress has been made in terms of analyzing these video for tasks like action classification~\cite{kay2017kinetics, kuehne2011hmdb,soomro2012ucf101}, action localization~\cite{caba2015activitynet,zhao2019hacs}, video description~\cite{xu2016msr,das2013thousand}, video question answering~\cite{lei2018tvqa,tapaswi2016movieqa,zeng2017leveraging}, object and scene understanding~\cite{zhao2019hacs}, etc. The majority of these tasks are focused on recognizing visual aspects present/happening in the video, e.g., action, scene, object detection, and classification. 

\begin{figure}[t]
    \centering
    \includegraphics[width =0.8\columnwidth]{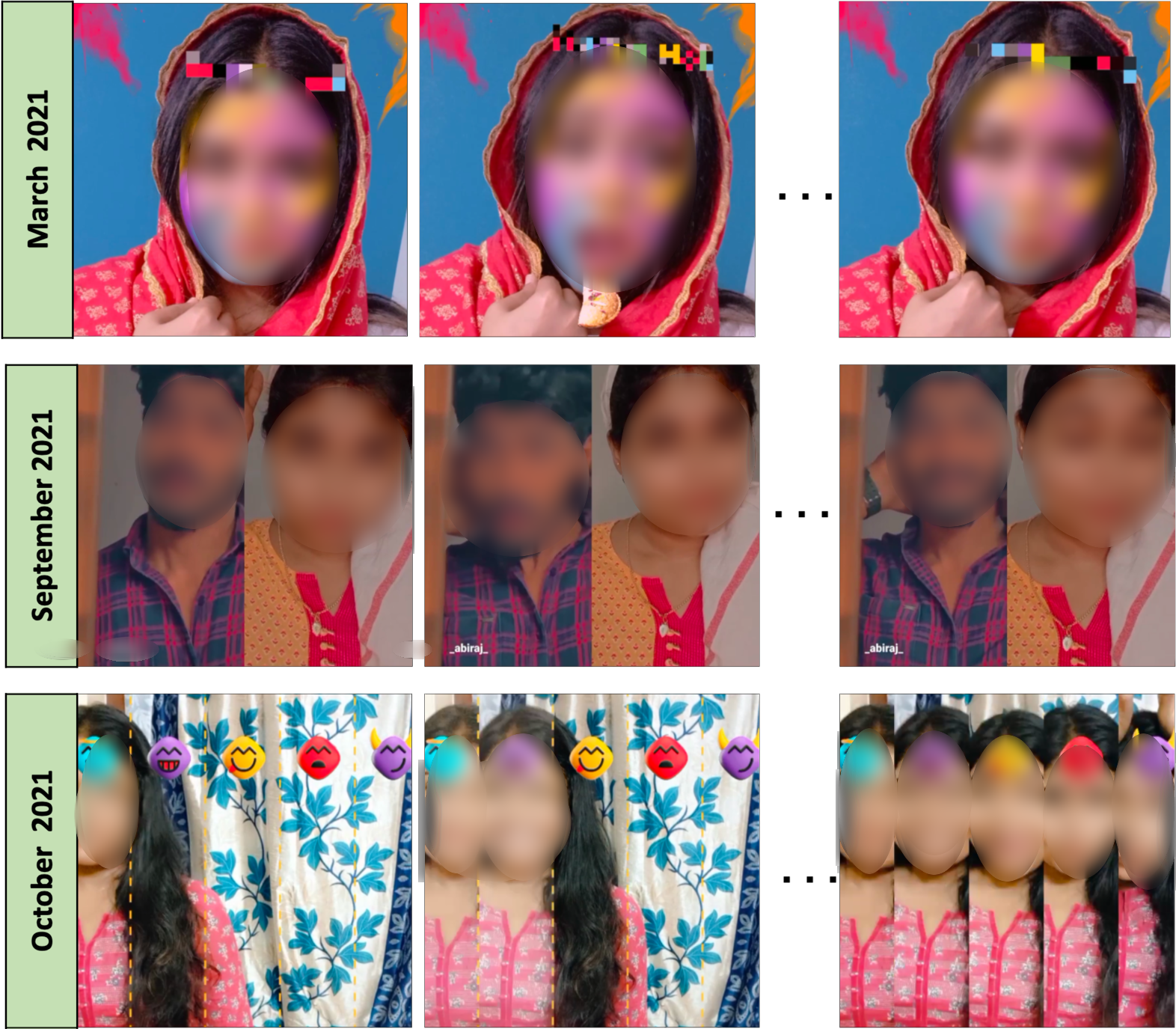}
    \caption{\small{\textbf{\datasetname: }We highlight three videos uploaded by a particular user. The concept labels for these are \textit{festival}, \textit{couple romance}, \textit{comedy} respectively. We also see the diversity in the video types, \textit{self-shot}, \textit{split-screen} and \textit{special effects}. We also observe how the content is temporally aligned to real-word events; for instance, the festivals. \datasetname has $50$K such annotated videos across $11$ languages with masked user identifiers and timestamps for deeper semantic analysis of social media content. \textit{Faces have been blurred for preserving privacy.}}}
    \label{fig:cover}
\end{figure}

Detecting these visual aspects helps in answering \textit{what occurs in a video?} But, it does not capture \textit{how viewers interpret the video?} and \textit{which concept(s) the creator of the video wishes to convey?} In this work, we investigate the semantic understanding of videos uploaded on short-video social media platform - \textit{Moj}~\footnote{https://mojapp.in} from the perspective of \textit{creators} and \textit{viewers} of these videos, which has not been explored before, primarily due to the lack of large-scale annotated video datasets. Considering the rapid adoption of social media, a holistic understanding of the creation, consumption, and popularity dynamics of these videos forms an important and timely research direction. 

To facilitate this under-explored research direction, we present a novel dataset, \datasetname, built from short videos posted on the short-video platform - \textit{Moj}. Even though existing datasets for semantic understanding source videos from social media (\eg YouTube~\cite{abu2016youtube}, Vine~\cite{nguyen2016open}, Facebook~\cite{ray2018scenes}), they are not suitable for our task. We highlight the key challenges and elaborate on how \datasetname addresses them:
    \begin{itemize}[noitemsep]
    \item \noindent\textbf{Taxonomy:} Prior datasets ~\cite{abu2016youtube, goyal2017something, kay2017kinetics} adopt a top-down approach of constructing a vocabulary of visual concepts from domain-independent taxonomies (\eg freebase) and mining videos from social media using this vocabulary. However, this vocabulary is not exhaustive and fine-grained for capturing popular concepts in social media discourse. Moreover, this method generates "easy videos" as search engines prioritize them first~\cite{ray2018scenes}.~\cite{ray2018scenes} adopt uniform sampling to address this problem while we construct a comprehensive bottom-up taxonomy using popularity-based sampling of videos for bridging this gap.
    
    \item \noindent\textbf{Novel video types:} Existing datasets do not capture novel and challenging video formats like split-screen videos, special effects (masks/graphics overlaid on faces), portrait videos, lip-syncing to pre-recorded audio, etc. (Figure~\ref{fig:cover}) which are dominant on social media platforms. \datasetname curates the videos from a short video platform - \textit{Moj} and annotates them for these media types for filling this gap.

    \item \noindent\textbf{Video Narrative:} Broadly speaking, there are three distinct kinds of videos on social media: a) \textbf{\textit{Micro Narrative}}: Videos which are short in duration~\cite{nguyen2016open} (5-6 secs) or are clipped out from longer videos~\cite{monfort2019moments, zhao2019hacs, diba2019holistic}, b) \textbf{\textit{Long Narrative}}: Longer videos ~\cite{abu2016youtube, yu2019activitynet, Damen2021PAMI}, usually more than 1-2 minutes, which tell a more detailed narrative or story c) \textbf{\textit{Short Narrative}}: These are longer than micro-videos (typically 10-20 secs) and provide authors and content creators more flexibility in terms of time limits. Despite the explosive growth of short video platforms like Tiktok, Reels, Youtube Shorts, and Moj, short videos have not been explored in detail in the Computer Vision and AI communities, primarily because of the lack of a large-scale labeled dataset.  \datasetname contains complete videos created with a short and concise narrative presenting an opportunity to understand this new avenue of video understanding.
    
    \item \noindent\textbf{Sparse/Noisy Hashtags:} Since expert annotation is expensive, large datasets often use hashtags added by the creators~\cite{nguyen2016open}. However, hashtags are usually sparse - 56\% of videos did not have hashtags in MV-58~\cite{nguyen2016open}. Also, they can be noisy, as shown in~(\ref{subsec:hashtag-analysis-appendix}). Our dataset, \datasetname, addresses this by manually annotating the videos using expert annotators.
    
    \item \noindent\textbf{Linguistic Diversity:} Existing datasets for semantic understanding of videos are not motivated towards exploring linguistic diversity while \datasetname comprises of videos from $11$ languages, annotated with the language of the audio for facilitating multilingual semantic understanding of videos.
    
  \end{itemize}

\datasetname contains concept, affective states, audio type, video type and language annotations for understanding the \textit{creator's} and \textit{viewer's} perspectives. We label the videos with the following annotations for modeling the \noindent\textbf {viewer's perspective:}

\begin{figure*}[t]
   \begin{subfigure}[h]{\textwidth}
   \centering
    \includegraphics[width=.8\linewidth]{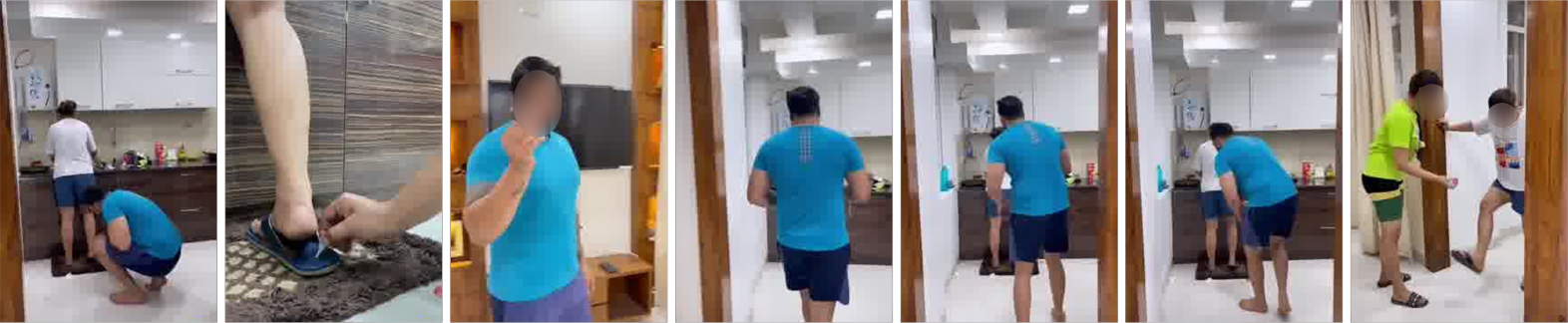}
    \caption{\textbf{Prank Scene}: A man is trying to prank the lady by putting an adhesive on her footwear with the intent of creating a funny situation for the viewers. Deep semantic understanding is required to understand the spatio-temporal-audio context of the scene to classify as "prank" because detection of visual or audio aspects is not sufficient.}
    \label{fig:prank}
    \end{subfigure}
 
    \begin{subfigure}[h]{\textwidth}
    \centering
    \includegraphics[width=.8\linewidth]{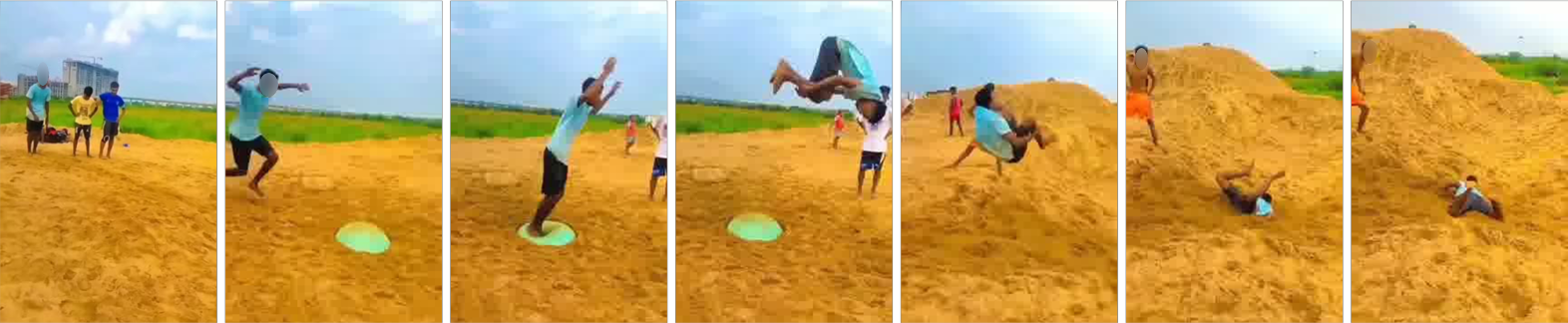}
    \caption{\textbf{Fail Scene}: Kid is trying to perform a summersault using a small trampoline but fails to complete the flip. For correct classification, model needs to focus on the unplanned fall at the end of the video to classify it as a "fail" video.}
    \label{fig:fail}
    \end{subfigure}

    \begin{subfigure}[h]{\textwidth}
    \centering
    \includegraphics[width=.8\linewidth]{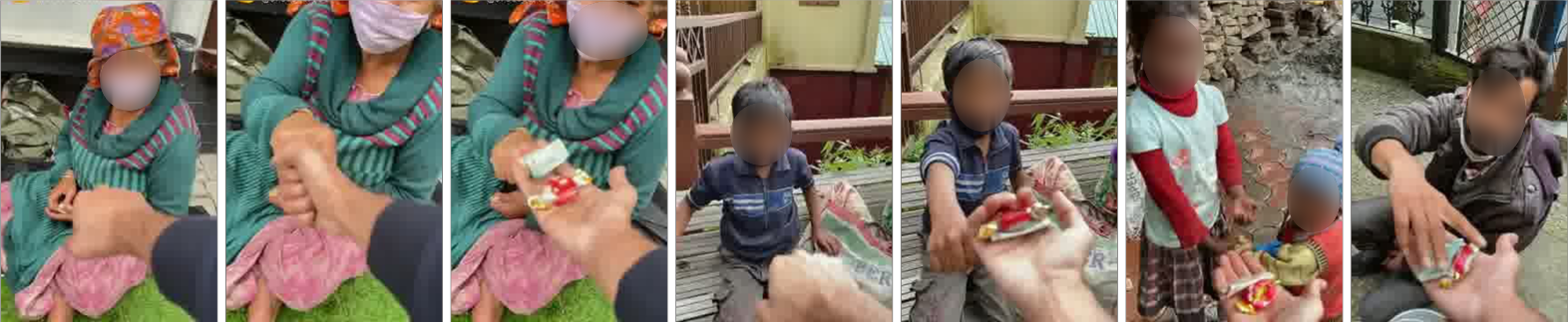}
    \caption{\textbf{Philanthropy Scene}: A man meets and greets needy strangers and surprises them with a gift. In order to recognize this as a gesture of kindness, our model needs to understand the economical situation and emotional state of the subjects in the videos and focus on the exchange of tokens.}
    \label{fig:kindness}
    \end{subfigure}

    \begin{subfigure}[h]{\textwidth}
    \centering
    \includegraphics[width=.8\linewidth]{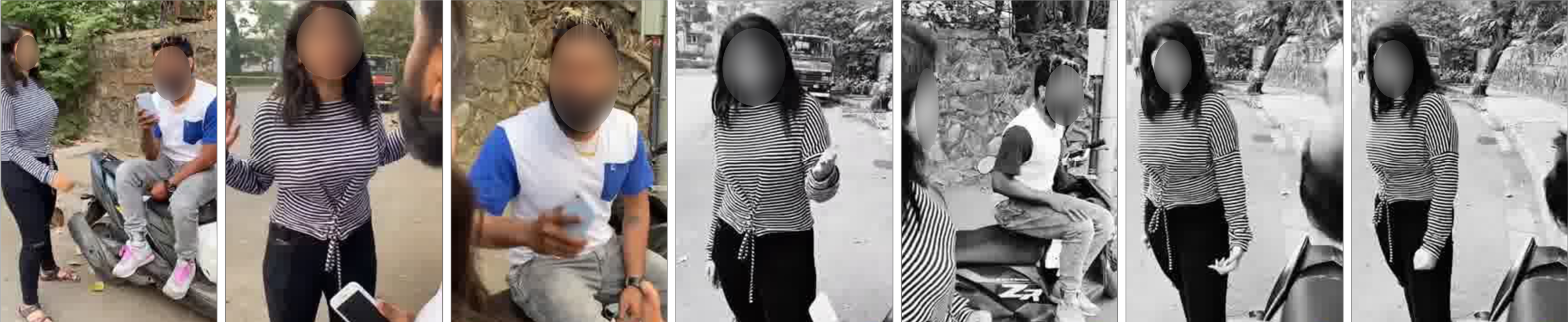}
    \caption{\textbf{Comedy Scene}: A funny and sarcastic verbal exchange between two friends. Both display a range of emotions during the act but the overall outcome of the video is a comedic situation. Focussing on facial emotions or human pose might not be sufficient for understanding the scene.}
    \label{fig:comedy}
    \end{subfigure}

\caption{\textbf{Unique Concepts present in \datasetname:} Our theme taxonomy comprises of several unique topics popular in social media domain but unexplored in literature: (a) Prank videos showing planned mischievous acts aimed to elicit reactions from co-creators \cite{Jarrar2019PerceptionOP}; (b) Fail videos that record unsuccesful attempts resulting in harm-joy \cite{Roseman2018ConcludingCS}; (c) Philanthropy videos portraying acts of helpful service, moral assistance or charitable deeds; (d) Scripted and natural comedy videos which can be further categorized based on the inter-agent relationships between the actors - couple, family, kids, friends, etc. \textit{Faces have been blurred for preserving privacy.}}
\label{fig:qual}
\end{figure*}

    \begin{itemize}[noitemsep]
         \item \textbf{Concept:} Each video is annotated for a concept~(across $34$ labels) by expert annotators. Our dataset contains widely popular and unique social media concepts like \textit{pranks, fails, romance, philanthropy, comedy, etc.} Figure~\ref{fig:qual} shows some examples which demonstrate that understanding these videos, which are very human-centric, self-shot with a short story goes beyond detecting and classifying the audio-visual aspects and makes \datasetname~challenging.
         \item \textbf{Affective States:} We provide annotations for $11$ emotion categories present in these videos.
    \end{itemize}
    Similarly, to understand the \textbf{creator's perspective}, we provide annotations for media types that content creators use to convey their point. Figure~\ref{fig:cover} shows some of the examples.
    \begin{itemize}[noitemsep]
          \item \textbf{Audio Types:} The audio types are unique and diverse with recorded/self-sung songs, dialogues, monologues, instrumentals, etc.  
          \item \textbf{Video Types:} Video formatting comprises of slideshows, animations, split-screens, self-shot, movie/TV-serial clips, etc. which are very popular on short video platforms. 
    \end{itemize}
    
Additionally, our dataset \datasetname~ can be used for various tasks and applications, such as: 

    \begin{itemize}[noitemsep]

        \item \textbf{Multilingual Modeling:} We provide annotations for the $11$ different languages, opening opportunities for multilingual semantic understanding.
        
        \item \textbf{Creator Modeling:} We also provide masked creator identifiers and recent videos uploaded by these creators (100k videos), opening up exciting user modeling ideas inspired by semantic video understanding. 
        
        \item \textbf{Temporal Analysis:} Social media content has a very short life span and is very dynamic. To enhance understanding here, we provide timestamps of these videos, which can help model temporal dynamics of the nature of popular content on such platforms. Moreover, we provide masked user profiles to identify videos from the same creators to analyze the shift in their perspectives over time. 
        
    \end{itemize}

To the best of our knowledge, \datasetname is the first human-annotated large-scale dataset of short videos that can be used for modeling concepts, affective states, and media types across $11$ languages, presenting a unique opportunity for understanding social media content. Overall, \datasetname contains 900 hours of video data uploaded by $23121$ creators with $50$K expertly annotated videos and $100$K unlabeled videos with an average duration of around 20 seconds. We also present baseline results to empirically establish that \datasetname is challenging and unique in Section ~\ref{sec:experiments}. In Section~\ref{sec:tasks}, we discuss the application of \datasetname over various research problems.
\begin{table*}[t]
    \centering
    \resizebox{\textwidth}{!}{
    \begin{threeparttable}
    \begin{tabular}{crcccccccccc}
    \toprule[1.5pt]
	&Datasets&	Size& 	Duration & Source&	Labels&	Audio Types&	Video Types&	Affective&	Focus & Lang & Year\\
	\midrule
	\multirow{1}{*}{\rotatebox{0}{Image}} & Intentonomy~\cite{jia2021intentonomy} &	14k & - & Flickr &	\cellcolor{green!25} HA	&\cellcolor{red!25}-&	\cellcolor{red!25}-& \cellcolor{red!25}NA &	Understanding Intent of Social Media Posts & \cellcolor{red!25} -  & $`20$\\
	
	\midrule
	
	& Sports-1M~\cite{karpathy2014large} &	1M & $4$ min &	YouTube&\cellcolor{red!25} 	MG&	\cellcolor{red!25}NA&	\cellcolor{red!25}NA&	\cellcolor{red!25}NA&	Sports Activity Classification (487 Classes)&\cellcolor{red!25}NA&$`14$\\
	
	&	ActivityNet~\cite{yu2019activitynet} & 27801& $5$-$10$ mins	 &	Web &\cellcolor{green!25} 	HA&	\cellcolor{red!25}NA&	\cellcolor{red!25}NA&	\cellcolor{red!25}NA&	HAR (203 classes) &\cellcolor{red!25} NA&$`15$\\
    
    &	MV-58K~\cite{nguyen2016open} & 260k & $6$ secs & Vine & \cellcolor{red!25} MG & \cellcolor{red!25}NA &	\cellcolor{red!25}NA	&\cellcolor{red!25}NA	&Activity, Objects, Platform Specific Classes&\cellcolor{red!25} NA&$`16$\\
	
	&Charades~\cite{sigurdsson2016hollywood} &	10k & $30$ secs &	CrowdSourced&\cellcolor{green!25} 	HA&	\cellcolor{red!25}NA&	\cellcolor{red!25}NA&	\cellcolor{red!25}NA&	HAR + Object Classification (157 classes)&\cellcolor{red!25}NA&$`16$\\

	\multirow{12}{*}{\rotatebox{0}{Video}}&YouTube-8M~\cite{abu2016youtube} &	8M&	$2$-$10$ mins & YouTube&\cellcolor{red!25} 	MG&\cellcolor{red!25}NA&	 \cellcolor{red!25}NA&	\cellcolor{red!25}NA&	Video Topic Classification&\cellcolor{red!25}NA &$`16$\\
	
	&	Kinetics~\cite{kay2017kinetics} &300k & $10$ secs	&	YouTube&\cellcolor{green!25} 	HA&	\cellcolor{red!25}NA&	\cellcolor{red!25}NA&	\cellcolor{red!25}NA&	HAR (400/600/700 classes)&\cellcolor{red!25} NA&$`17$\\
    
	& Something-Something~\cite{goyal2017something} & 100k & $2$-$6$ secs & CrowdSourced &\cellcolor{green!25} HA &	\cellcolor{red!25}NA &	\cellcolor{red!25}NA	& \cellcolor{red!25}NA	& HAR (174 classes) & \cellcolor{red!25} NA&$`17$\\
	
    & Epic-Kitchens~\cite{Damen2021PAMI} &	39594 &	$1$-$55$ mins & CrowdSourced	&\cellcolor{green!25} HA	&\cellcolor{red!25}NA&	\cellcolor{red!25}NA&	\cellcolor{red!25}NA&	Actions in Kitchen &\cellcolor{green!25} Yes&$`18$\\
	
	& SOA~\cite{ray2018scenes} & 562k & $10$ secs & Facebook & \cellcolor{green!25} HA & \cellcolor{red!25}NA &	\cellcolor{red!25}NA	&\cellcolor{red!25}NA	&Scenes, Objects, Actions &\cellcolor{red!25} NA& $`18$\\
    
    &	MomentsInTime~\cite{monfort2019moments} &1M & $3$ secs &	10 sources &\cellcolor{green!25} 	HA&	\cellcolor{red!25}NA&	\cellcolor{red!25}NA&	\cellcolor{red!25}NA&	339 action classes &\cellcolor{red!25} NA&$`19$\\

    & HACS~\cite{zhao2019hacs} &	1.5M& $2$ secs &	YouTube&\cellcolor{green!25} 	HA&	\cellcolor{red!25}NA&	\cellcolor{red!25}NA&	\cellcolor{red!25}NA&	HAR (200 classes)&\cellcolor{red!25} NA&$`19$\\
    
	& HVU~\cite{diba2019holistic} &	500k & $\leq10$ secs & YT8M, Kinetics, HACS	&\cellcolor{red!25} MG	&\cellcolor{red!25}NA&	\cellcolor{red!25}NA&	\cellcolor{red!25}NA&	Actions, Objects, Concepts, Events, Attributes, Scenes&\cellcolor{red!25} NA & $`20$\\
	
	\midrule
	\multirow{2}{*}{\rotatebox{0}{Affects (videos)}}
	
	&	CMU-MOSEI~\cite{zadeh2018multimodal} &23K& $7.3$ secs &		YouTube&\cellcolor{green!25} 	HA&\cellcolor{red!25}NA&	\cellcolor{red!25}NA&	\cellcolor{green!25}Yes&	Perceived EC (6 classes)&\cellcolor{red!25} NA&$`18$\\
	
	&EEV~\cite{sun2020eev} & 23k & $1$-$25$ mins & Online Videos &\cellcolor{red!25} MG &	\cellcolor{red!25}NA & 	\cellcolor{red!25}NA&	\cellcolor{green!25}Yes&	Evoked EC (15 classes)&\cellcolor{red!25}NA&$`20$\\
	
	\midrule

	\multirow{2}{*}{}&\multirow{2}{*}{\textbf{Ours}} &	\multirow{2}{*}{50k(+100k)} &- & \multirow{2}{*}{Social Media Platform}	&	\cellcolor{green!25}&\cellcolor{green!25}Annotated for&\cellcolor{green!25} Annotated for&	\cellcolor{green!25}&\multirow{2}{*}{Concept, Affective States, Media Type, Language}&\cellcolor{green!25}Annotated for & $`21$ \\	
	
    &&&&&  \cellcolor{green!25}\raisebox{2.7ex}[2.7ex]{\multirow{2}{*}{HA}} &\cellcolor{green!25}7 classes&\cellcolor{green!25}8 classes&\cellcolor{green!25}\raisebox{2.7ex}[2.7ex]{\multirow{2}{*}{Yes}}&&\cellcolor{green!25}11 languages \\

    
	\bottomrule[1.5pt]
    \end{tabular}
    \end{threeparttable}
    }
    \begin{tablenotes}\footnotesize
    \item $^\dagger$ NA, MG, HA indicate ``not annotated'', ``machine generated'', and ``human annotated'', respectively.
    \end{tablenotes}
    
    \caption{Comparison of \datasetname with related image and video datasets. Our dataset has exhaustive and expertly annotated annotations for concepts, audio/video types, affective states and audio language for social media short videos. Majority of the other datasets focus on specific tasks like action classification and do not annotate for other dimensions. YT8M, SOA and HVU adopt more holistic annotations. We report the range or average duration of videos for the datasets.}
    \label{tab:datasets}
\end{table*}

\section{Related Work}
\label{sec:related}
We review related datasets for semantic understanding of videos from social media and summarise them in Table~\ref{tab:datasets}.
\subsection{Semantic Understanding Datasets}
Various datasets and tasks have been proposed for video understanding. \\
\textbf{Action classification} is a popular research problem for which benchmark datasets like~\cite{kay2017kinetics, carreira2019short, Goyal_2017_ICCV, monfort2019moments, kuehne2011hmdb, soomro2012ucf101, Kong_2019_ICCV, caba2015activitynet, Damen2021PAMI, karpathy2014large, zhao2019hacs, li2020ava} have been proposed.

\noindent\textbf{Concept Understanding:} Going beyond action classification, detection, and segmentation of visual elements, \textit{theme/concept} classification datasets focus on modeling interplay between the visual and audio elements for understanding the overall theme/concept represented by the videos. For instance, YouTube-8M~\cite{abu2016youtube} focuses on classifying videos into categories like \textit{fashion, games, shopping, animals, etc.}. The taxonomy has been curated manually to capture purely \textit{visual} categories, and the dataset has been machine annotated using the YouTube Video Annotation system for collecting videos. Similarly, Holistic Video Understanding (HVU)~\cite{diba2019holistic} annotate videos from ~\cite{abu2016youtube, kay2017kinetics, zhao2019hacs} for concepts along with scenes, objects, actions, attributes, and events using Google Vision API and Sensifai Video Tagging API. \textbf{MicroVideos}~\cite{nguyen2016open} contributes videos collected from a micro-video application - Vine and interpret user-generated hashtags as annotations. More recently, datasets for understanding \textbf{Intent} and \textbf{Motivation} from social media posts are being investigated~\cite{jia2021intentonomy,siddiquie2015exploiting, kruk2019integrating,wang2019persuasion,Vondrick_2016_CVPR,zhang2018equal, ye2019interpreting}.

\noindent\textbf{Other Video Understanding Tasks:}~\cite{russakovsky2015imagenet, real2017youtube, perazzi2016benchmark, wang2014cdnet, dave2020tao, milan2016mot16} have been proposed for object detection, segmentation and tracking from videos. At the intersection of vision and language, datasets for video description~\cite{wang2019vatex,xu2016msr}, question-answering~\cite{lei2018tvqa,zeng2017leveraging,tapaswi2016movieqa}, video-object grounding~\cite{chen2019weakly,zhang2020does} and text-to-video retrieval~\cite{lei2020tvr,anne2017localizing} have been proposed. SVD~\cite{jiang2019svd} contribute a dataset for near-duplicate video retrieval.

\subsection{Affective Analysis of Social Media Content}
Understanding perceived emotions of individuals using verbal and non-verbal cues is an important problem in both AI and psychology for various applications. One such application is for understanding the projected~\cite{zadeh2018multimodal} and evoked emotions~\cite{kassam2010assessment, micu2010measurable} from multimedia content like advertisements and movies. There is vast literature in inference of perceived emotions from a single modality or a combination of multiple modalities like facial expressions~\cite{face1,face2}, speech/audio signals~\cite{speech1}, body pose~\cite{body}, walking styles~\cite{step} and physiological features~\cite{physiological1}. There has been a shift in the paradigm, where researchers have tried to fuse multiple modalities to perform emotion recognition, also known as Multimodal Emotion Recognition. Fusion methods like early fusion~\cite{early-fusion}, late fusion~\cite{late-fusion}, and hybrid fusion~\cite{hybrid-fusion} have been explored for emotion recognition from multiple modalities.
\subsection{Research Problems with Social Media Content}
\noindent \textbf{Multilingual Analysis of Videos: }Multilingual analysis of images and videos has been studied previously. Harwath et al.~\cite{harwath2018vision} proposed a bilingual dataset comprising English and Hindi captions. Ohishi et al.~\cite{ohishi2020trilingual} extended this dataset to include Japanese captions and proposed a trilingual dataset. Approaches for bilingual video understanding include~\cite{kamper2018visually, azuh2019towards, ohishi2020pair}. On the other hand, several datasets for multilingual video understanding~\cite{sanabria2018how2, wang2019vatex} along with techniques for analyzing them~\cite{rouditchenko2021cascaded} have been proposed, although they lack diversity in audio language.\\
\noindent \textbf{User Modeling of Social Media Content: } People are increasingly relying on social media platforms for sharing their daily lives, which reflect their personality traits and behavior. User modelling based on their online persona and activity has been successfully leveraged for digital marketing~\cite{yogesh2019digital, alalwan2017social} and content recommendation~\cite{yuan2020parameter,wu2020diffnet++}. Not only on the consumer side, but user profiling is also helpful for helping content creators on such social media platforms~\cite{arriagada2020you, huotari2015analysis}. To further research in these directions, we provide masked user identifications. \\
\noindent \textbf{Temporal Analysis of Social Media Content: } A unique characteristic of social media content is the short life span of posts~\cite{fiebert2014life}. Such dynamically and temporally evolving content is evident and can be mapped to major festivals, celebrations, political events, news, and trends~\cite{haralabopoulos2015lifespan}. Such dynamic and temporally evolving content can be helpful to understand social media platforms better. 
\section{Our Dataset: \datasetname}
\label{sec:dataset}
In this section, we introduce \datasetname~and elaborate on the dataset collection and annotation process. 
\begin{figure*}[t]
\centering
   \begin{subfigure}[h]{.78\textwidth}
\includegraphics[width=\linewidth]{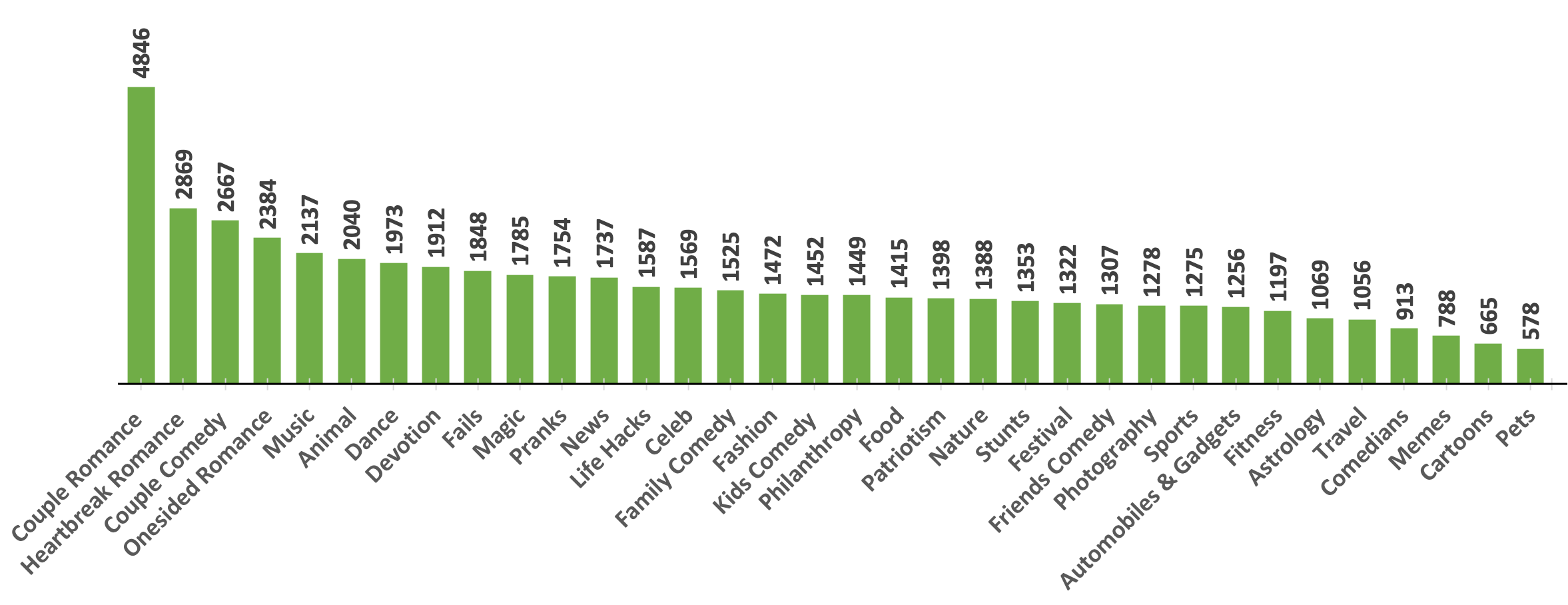}
    \caption{Concept Taxonomy}
    \label{fig:theme-distribution}
  \end{subfigure}
 \begin{subfigure}[h]{.23\textwidth}
  \centering
    \includegraphics[width=\linewidth]{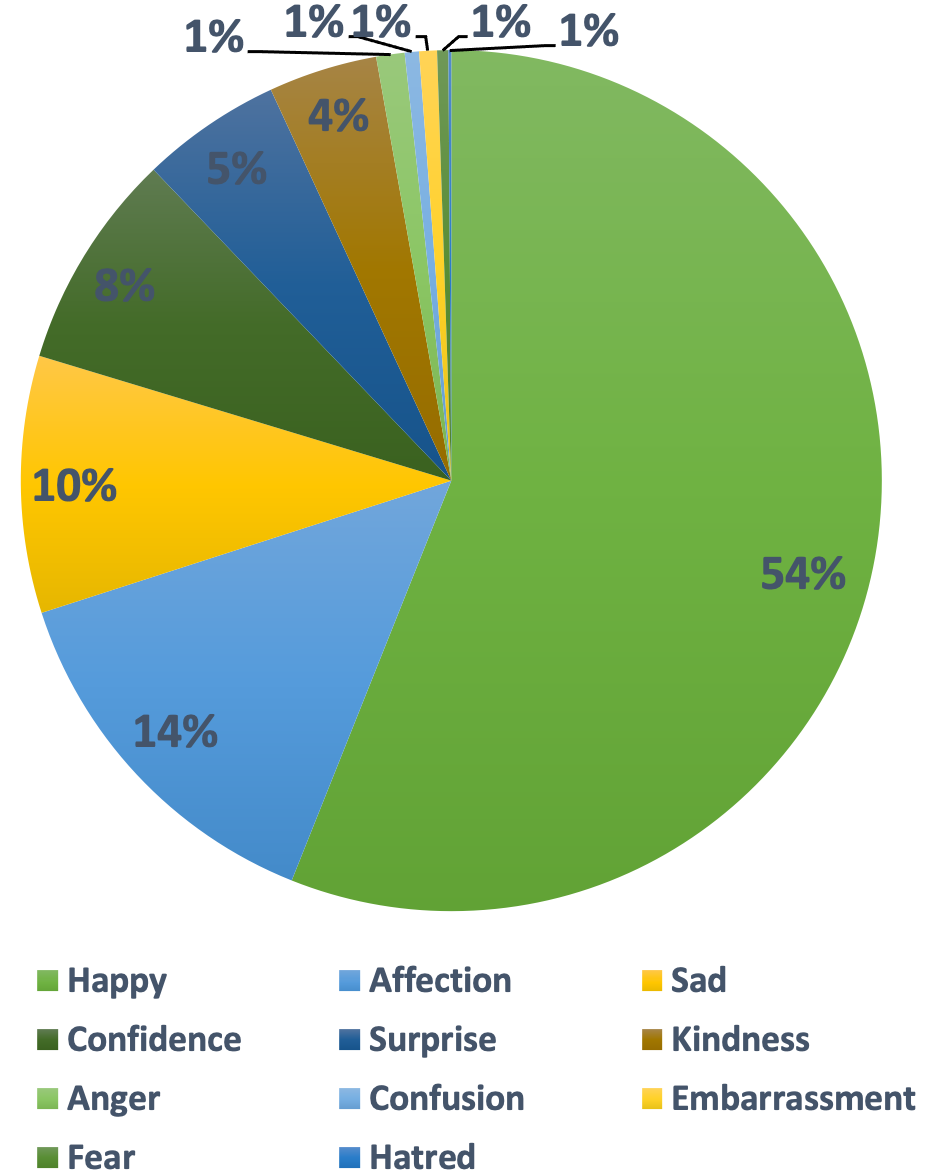}
    \caption{Affective Labels}
    \label{fig:affective-distribution}
  \end{subfigure}
    \begin{subfigure}[h]{.23\textwidth}
   \centering
    \includegraphics[width=\linewidth]{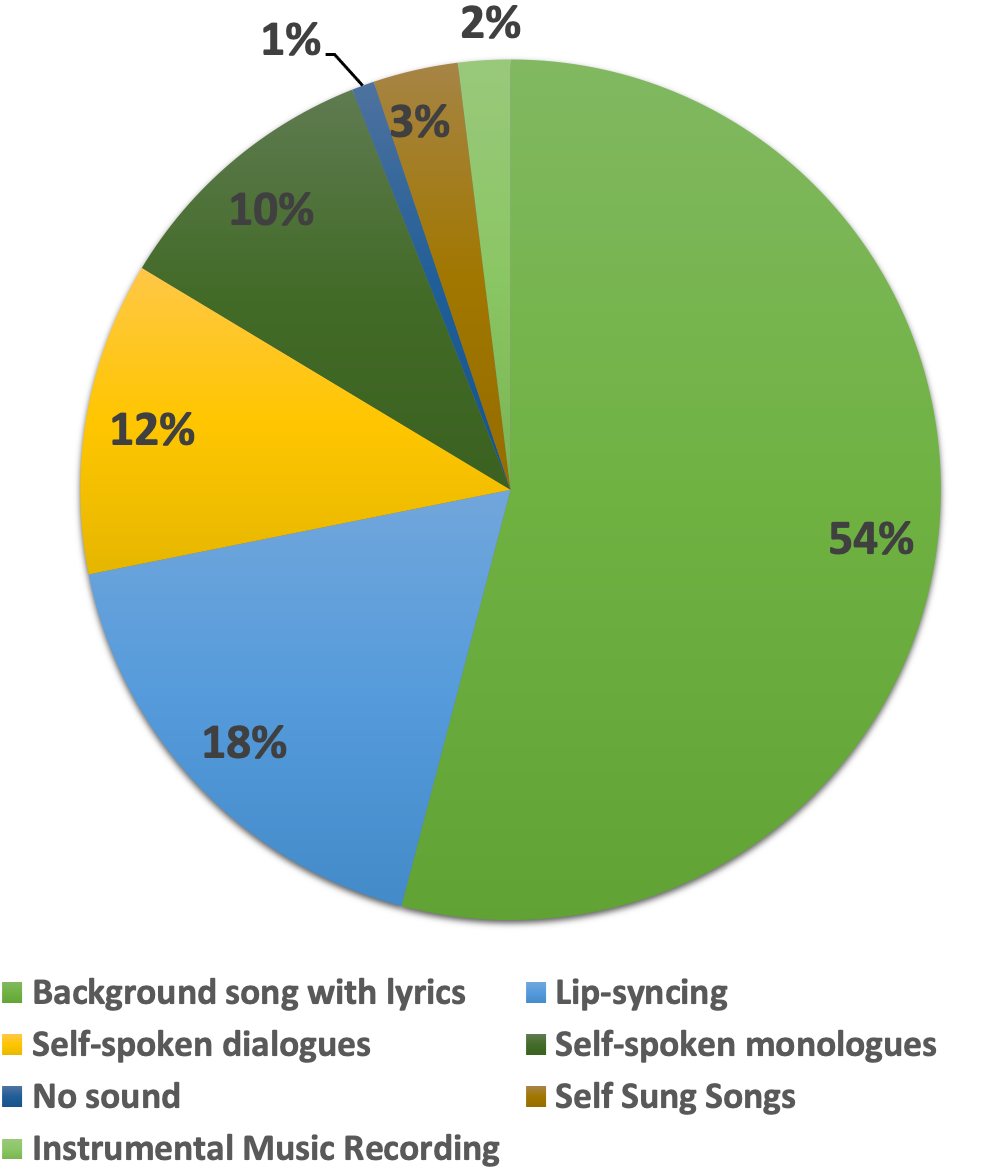}
    \caption{Audio Type}
    \label{fig:audio-type-distribution}
  \end{subfigure}
\begin{subfigure}[h]{.23\textwidth}
    \includegraphics[width=\linewidth]{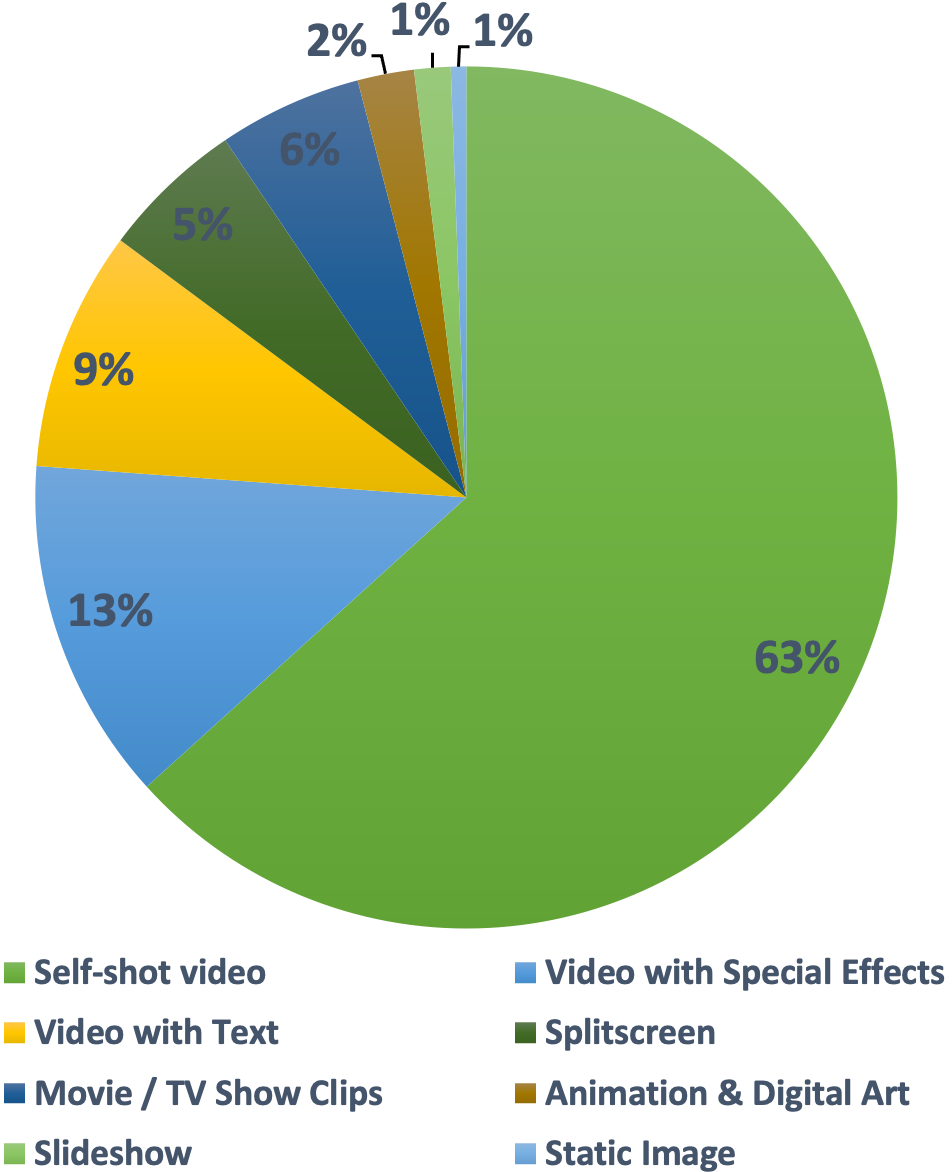}
    \caption{Video Type}
    \label{fig:video-type-distribution}
  \end{subfigure}
\begin{subfigure}[h]{.23\textwidth}
    \includegraphics[width=\linewidth]{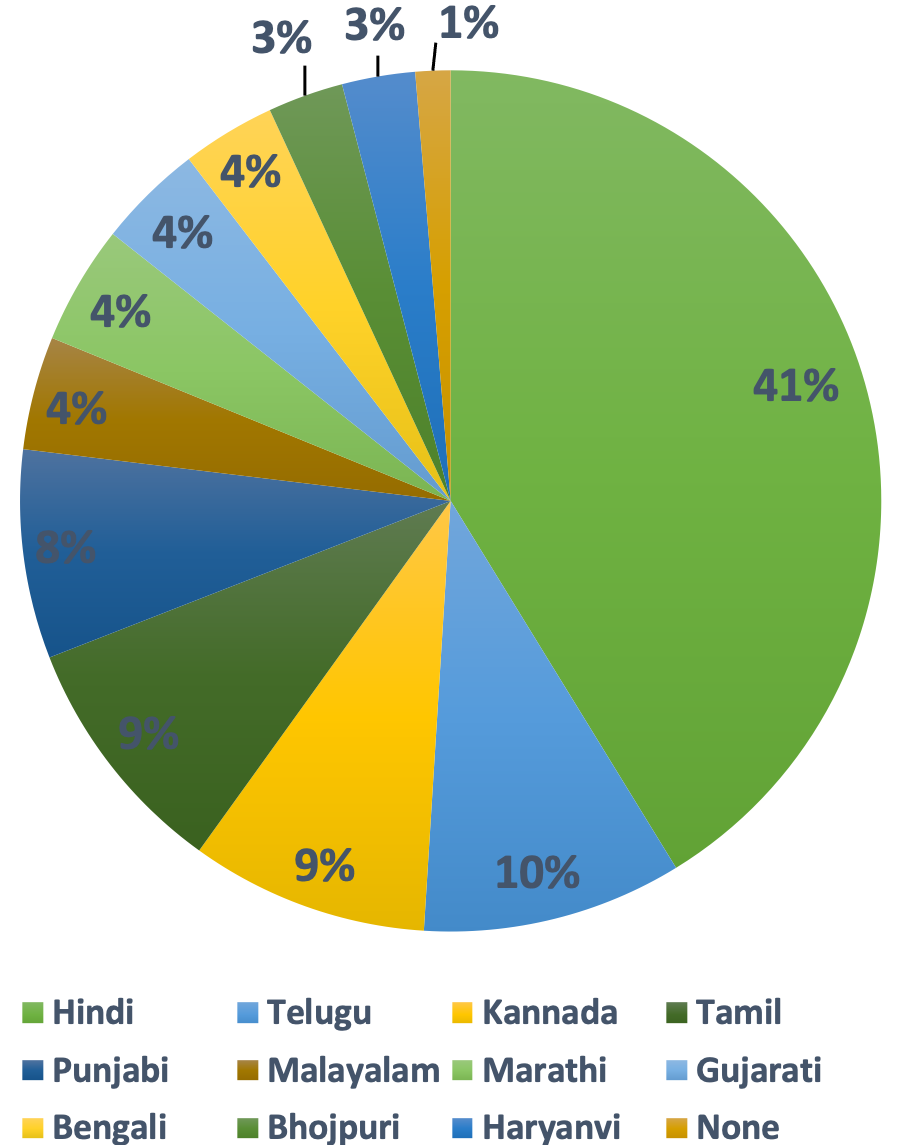}
    \caption{Language}
    \label{fig:language-distribution}
  \end{subfigure}
\caption{\textbf{ \datasetname Taxonomy:} Sub-figures ~\ref{fig:theme-distribution} -- \ref{fig:language-distribution}  show the taxonomy and label distributions in the proposed \datasetname dataset for concept, affective states, audio type, video type and language in anti-clockwise direction.}
  \label{fig:plots}
\end{figure*}

\subsection{Taxonomy}
\label{subsec:taxonomy}
\noindent We annotate our datatset for the following taxonomies. A detailed description of all the annotation labels of the taxonomy is presented in Appendix~\ref{subsec:annotation-label-description}. 

\noindent\textbf{Concept:} Creation of a taxonomy for concepts is a non-trivial exercise, requiring both comprehensiveness as well as frequency coverage. We adopted a bottom-up approach to model social media behavior rather than mining videos for an existing taxonomy. To achieve this, we employed a team of digital social media experts for scanning 1.5 million popular posts and assigned a label that concisely describes a post. The taxonomy grew to more than 1000 concepts and was pruned to $34$ popular labels covering more than 75\% of the videos for this study. Some of these concepts like \textit{fails, pranks, comedy, romance, philanthropy} are unique to our dataset and are illustrated in Figure~\ref{fig:qual}. We illustrate the distribution across these concepts in Figure~\ref{fig:theme-distribution}.

\noindent\textbf{Affective States:} We provide annotations for the projected affective labels for the videos. Inspired by \cite{de2012happy}, we adopt a $11$ label taxonomy for affective states. We present the distribution across these affective states in Figure~\ref{fig:affective-distribution}. 

\noindent\textbf{Audio Type:} Social media creators use a variety of audio styles like lip-syncing to pre-recorded songs, monologues, dialogues, self-sung songs, or instrumental music. We present a taxonomy of $7$ labels to cover the broad spectrum of audio content type~(Figure~\ref{fig:audio-type-distribution}). 

\noindent\textbf{Video Type:} We provide annotations for classifying video types based on how the video was created/edited~(Figure~\ref{fig:video-type-distribution}). The videos can be conventionally sourced from Movie or TV-Show clips or be self-shot on personal hand-held devices. The videos also contain slideshows, still images, and split screens. Additionally, many creators also publish videos with text to add a linguistic message to enhance the audio-visual effect. 

\noindent\textbf{Language:} We annotate audio language for our videos and highlight the linguistic diversity of our dataset in Figure~\ref{fig:language-distribution}.
\subsection{Data Collection}
\label{subsec:collection}
We collect our dataset from a leading short video application supporting over $15$ languages. The platform contains short videos uploaded by professional and amateur content creators on which users can view, like, share and comment. We extracted more than $1.5$M videos uploaded over $9$ months (Feb, $2021$ to Oct, $2021$) across $11$ languages and share $50k$ labeled and $100k$ unlabeled from this set. These videos were shortlisted based on platform engagement metrics after removing near-duplicates. The duration of videos ranges between $4.5 - 116$ seconds~(averaging $20$ seconds). Videos reported to be of sensitive nature and those containing nudity, violence, and abuse were removed. Additional steps about data collection are mentioned in~\ref{subsec:dataset-cleaning}.
\subsection{Data Annotation}
\label{subsec:annotation}
We employed domain experts in the field of social media who provided labels for the $50$K videos. Annotators were selected to ensure that we can label every video, across $11$ different languages, by experts who are fluent writers and speakers of the dominant language of the video. The annotators were provided with guidelines, which comprised of instructions about each task, definitions of class labels~(Appendix~\ref{subsec:in-depth-annotator-agreement}, Table~\ref{tab:label-taxonomy-description-big-table}) and a few worked-out examples to familiarize them with the annotation task.\\
\noindent \textbf{Annotator Onboarding:} We followed a strict annotator onboarding mechanism. We provided new candidates with a set of 100 posts that have been pre-annotated by expert reviewers and benchmarked against other candidates. Candidates not adhering to the benchmarks were not allocated further posts, and their responses were discarded.\\
\noindent \textbf {Inter-Annotator Agreement:} We evaluated inter-annotator agreements across all labels in different concepts using Krippendorff’s alpha (K-alpha)~\cite{Krippendorff2011ComputingKA} to account for labeling reliability amongst multiple annotators. All annotations were performed by $3$ annotators each, and their majority vote was accepted as the ground truth label. In case of a three-way disagreement, an expert annotator resolved the conflict and assigned the final label. 
The K-alpha values for the $4$ taxonomies, concept, audio type, video type, and affective states are $0.77$, $0.59$, $0.62$, and $0.40$, respectively. We present detailed per-label annotator agreement in Table~\ref{tab:annotator-agreement-big-table}. We observe strong agreements for most of the tasks. \datasetname~is finally split into train, validation, and test sets in a ratio of $60:20:20$.
\begin{table}[t]
\centering
\resizebox{.7\columnwidth}{!}
{
\begin{tabular}{lc} 
\toprule
\textbf{Data Description} & \textbf{Value}\\
\midrule
\# Concept & 34 \\
\# Languages & 11 \\
\# Affective States & 11 \\
\# Audio Types & 7 \\
\# Video Types & 8 \\
\# Creators & 23121 \\ 
\# Annotators & 95 \\ 
\# Labelled Videos & 55262 \\
\# Unlabelled Videos & 100K \\
\midrule
Total Duration Labelled & 310 hours \\
Total Duration Unlabelled & 600 hours \\
Average Duration & 20.2 $(\pm 9.5)$ seconds  \\
Min/Max Duration & 4.5/116 seconds\\
\bottomrule
\end{tabular}
}
\caption{\small{{\textbf{\datasetname Statistics}}}}
\label{table:dataset_summary}
\end{table}
\subsection{Dataset Analysis}
\label{subsec:analysis}
\datasetname contains 50K annotated and $100$K unlabeled videos with a total of $910$ hours of data. Figure~\ref{fig:theme-distribution} -- \ref{fig:language-distribution} show the exhaustive taxonomy and distribution of \datasetname. 

\noindent\textbf{Concept:} As evident from Figure \ref{fig:theme-distribution}, \textit{comedy} and \textit{romance} have a higher frequency than other labels, while \textit{pets} has the least frequency. This is expected given the trends in short video social media platforms that incentivize creators to create content with wide appeal. 

\noindent\textbf{Affective States:} Figure \ref{fig:affective-distribution} shows the $11$ affective states found in the corpus. We observe class imbalance that mirrors the distribution of natural human emotions. 

\noindent\textbf{Audio Type:} Figure~\ref{fig:audio-type-distribution} highlights an interesting phenomenon wherein more than $50$\% of the videos borrow the background music from a pre-recorded source while self-spoken dialogues and monologues are comparatively less. This alludes to the fact that a large majority of creators are more comfortable in visual mode of expression. Similarly, lip-syncing to existing audio is the second-most popular way of video creation. 

\noindent\textbf{Video Type:} As evident in Figure~\ref{fig:video-type-distribution}, more than two-third of videos sampled in the dataset are self-shot. Advances in photography have aided creators in adding visual as well as textual effects to the videos, making them the next most popular video formats. 

\noindent\textbf{Languages:} The dataset comprises videos in $11$ languages with Hindi as the majority language. 

\noindent\textbf{Duration:} \datasetname comprises of videos ranging from $4.5$s-$116$s with an average duration of $20$ seconds.  

\noindent\textbf{Creators:} \datasetname comprises of videos from $23121$ unique creators. A large majority of these creators ($15998$) contribute only one video in our dataset, while $7133$ contributed more than one video. This demonstrates the immense diversity of our dataset in terms of creators.

\noindent\textbf{Taxonomy Correlation:} 
In Appendix~\ref{subsec:more-data-analysis}, Figure~\ref{fig:correlation-heatmap-additional}, we present the correlation between concepts and affective states/media types. We observe that \textit{heartbreak romance} videos predominantly have \textit{sad} affective state; \textit{philanthropy} is strongly linked with \textit{kindness}. Similarly, we observe that videos with \textit{magic} label are linked with \textit{surprise} affective state; \textit{couple romance} shows the strongest predisposition towards \textit{affection}. These correlations provide insights that \datasetname comprises of videos that depict strong correlation with other underlying aspects and this correlation can be leveraged for better semantic understanding.

\section{Baseline Experiments}
\label{sec:experiments}
We perform baseline experiments to highlight the unique and challenging aspects of \datasetname.
\subsection{Concept Classification}
\label{subsec:baseline-concept-classification}
We report the results for concept classification using different modalities individually and in combination using late fusion in Table~\ref{tab:topic_classification}. We report top-1, top-3, and top-5 accuracy for all the experiments. 

\noindent \textbf{Audio-Visual Representation:} We experiment with 3D ResNet~\cite{hara2018can} backbones trained over Kinetics700~\cite{carreira2019short} for spatio-temporal modelling. We also evaluate deeper (R3D-101) and depth-wise separable architecture (R(2+1)D-50)~\cite{tran2018closer} but did not observe gains. Hence we use R3D-50 for all our experiments. For audio modelling, we leverage pretrained VGG~\cite{hershey2017cnn} model and CLSRIL23~\cite{gupta2021clsril}. VGG is trained for sound classification~(\cite{gemmeke2017audio}) and CLSRIL23 is trained over speech data of 23 Indic languages. We freeze the audio-visual backbones and train the classifier and multimodal fusion layers.

\noindent \textbf{Results and Discussion:} From Table~\ref{tab:topic_classification}, we observe that the performance of visual modality is higher than audio, which highlights the importance of visual modality for our dataset~\datasetname. On combining the modalities using late-fusion, we observe a gain of \textbf{$4\%$} (Row $6$ and $7$). This demonstrates the multimodal nature of the dataset. By combining both VGG and CLSRIL23 features with visual modality, we notice further gains showing complementary information in both these audio representations~(Row $8$). This is not surprising because our dataset contains a wide variety of audio types like \textit{songs, monologues, and dialogues.} While VGG has been trained for modeling sounds (music, vehicle, creek, instrument, etc.), CLSRIL23 is more specialized for understanding human speech. We expand on the training details and hyperparameters in Appendix~\ref{subsec:theme-baseline-additional-features}.  

\begin{figure}
     \centering
     \begin{subfigure}[b]{0.48\columnwidth}
         \centering
         \includegraphics[width=\columnwidth]{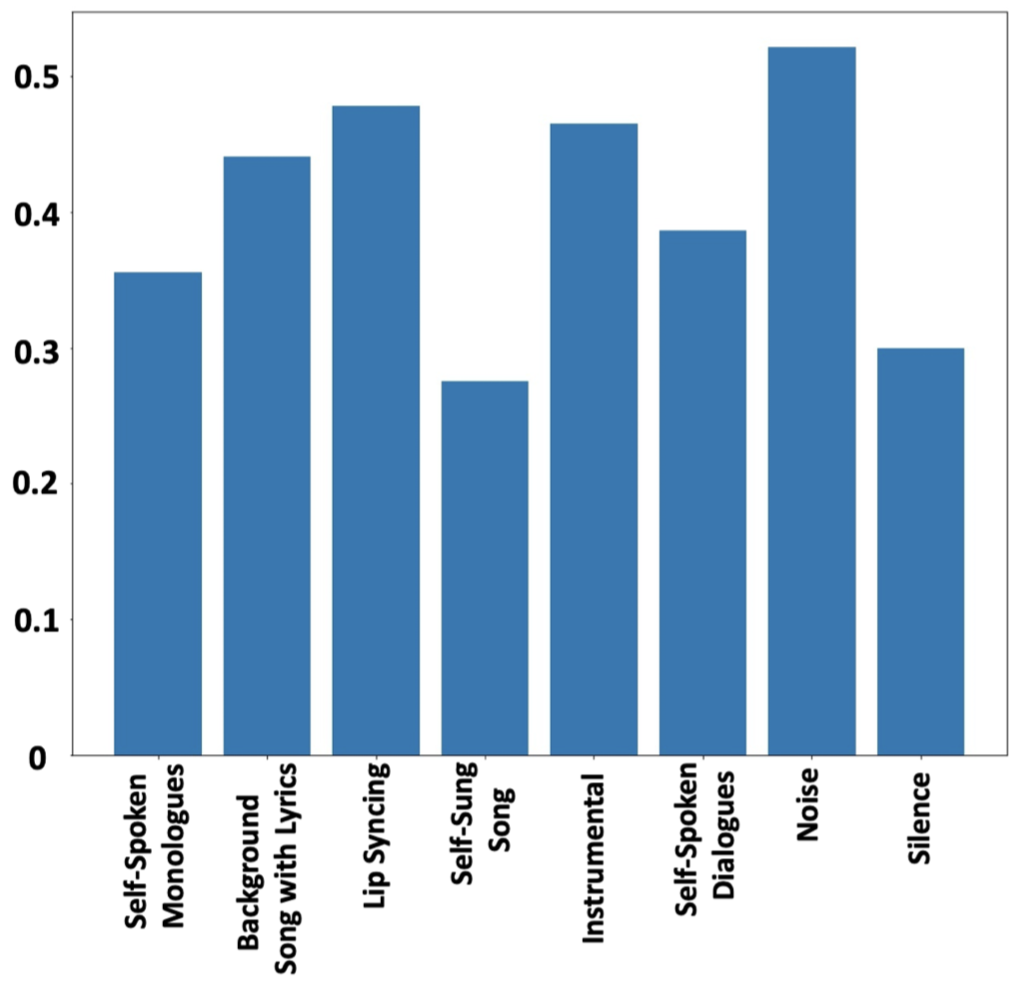}
         \caption{Error Rate in Concept Classification per Audio Type label}
         \label{fig:audio_err_dist}
     \end{subfigure}
     \hfill
     \begin{subfigure}[b]{0.48\columnwidth}
         \centering
         \includegraphics[width=\columnwidth]{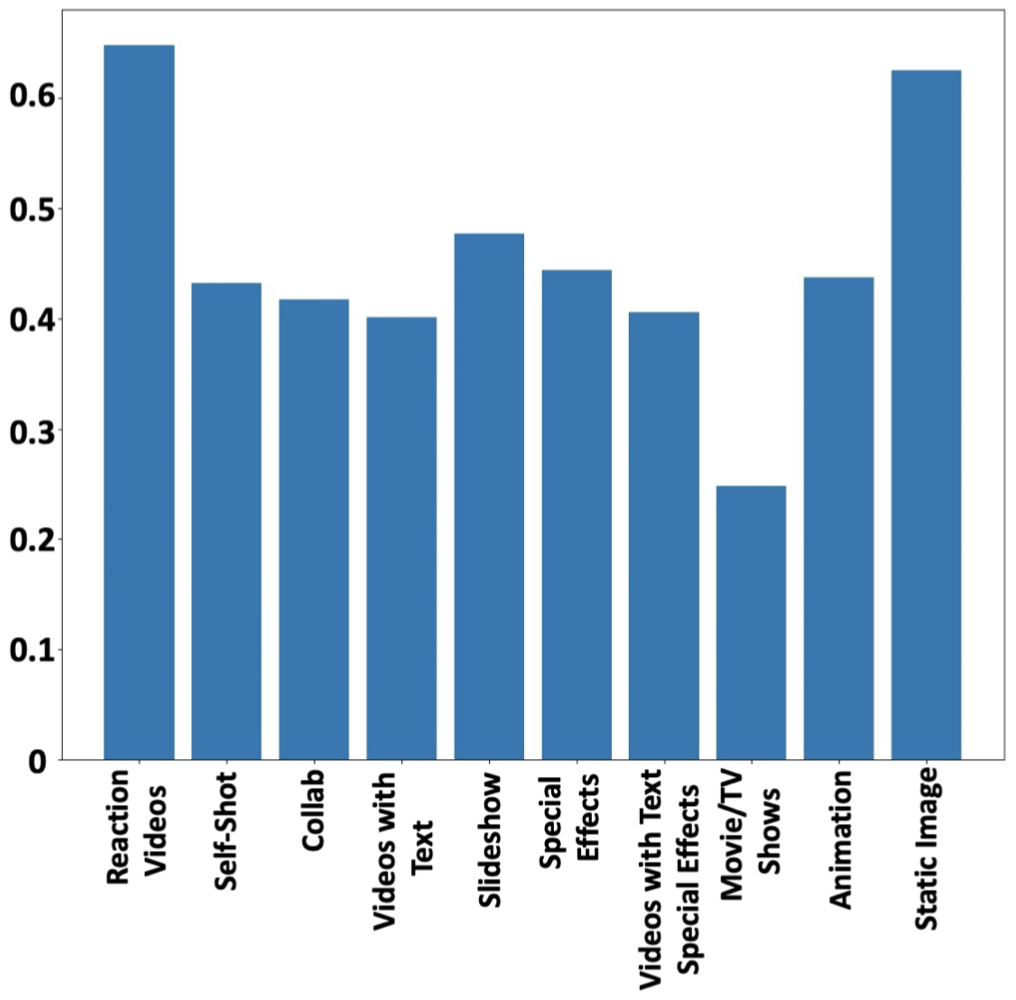}
         \caption{Error Rate in Concept Classification per Video Type label}
         \label{fig:video_err_dist}
     \end{subfigure}
      \caption{\textbf{Challenging Audio and Video Types:} We present an in-depth analysis into the misclassifications for concept classification. We try to understand the relation between the audio type and the video type of the incorrectly classified videos. }
     \label{fig:error_analysis}
\end{figure}

\noindent \textbf{Error Analysis:}  We analyze error cases for different media types in Figure~\ref{fig:video_err_dist} and Figure~\ref{fig:audio_err_dist}. We notice comparatively less performance on \textit{images, reaction videos, and slide-shows}, which showcases the novelty of these types in video datasets. \textit{Reaction videos} contain split-screens and are complex as the model needs to focus on the salient parts. Similarly, slide shows contain a lot of abrupt scene changes making it extremely challenging. On audio-types, we notice the model shows less accuracy for classes like \textit{lip-sync, instrumental, and silence/noise}. This is not unexpected as these do not provide relevant signals about the concept. Similarly, \textit{lip-sync} encodes the majority of the semantic information in the audio channel. These observations strongly highlight the unique challenges of our dataset \datasetname, which have not been explored before. In Figure~\ref{fig:confusion_matrix}~(in Appendix~\ref{subsec:theme-baseline-additional-analysis}), we plot the confusion matrix of the audio-visual model. We notice confusion among the concept labels like \textit{memes, kids, family, friends}, and \textit{couple comedy}, demonstrating the challenges in semantic understanding of such content.  We also study the impact on accuracy of concept categories using the audio-visual modalities in Figure~\ref{fig:audio_visual_diff}~(Appendix~\ref{subsec:theme-baseline-additional-analysis}). 
\subsection{Affective State Classification}
\label{subsec:baseline-emotion-classification}
We select two state-of-the-art affective state classification models and benchmark them on \datasetname. The results are summarized in Table~\ref{tab:emotion_classification}. We report top-1, top-3 and F1 scores. The first method, Kosti et al.~\cite{kosti2017emotic} is an emotion recognition model which uses the facial expressions of the dominant subject in the video and the background context. Tsai et al.~\cite{tsai2019multimodal} is a multimodal transformer-based model that uses both visual and audio modalities and has shown high performance on other emotion recognition datasets. We observe that the performance of these models on \datasetname is not very high. On further analysis of these models, we notice that videos associated with human-centric concept labels \textit{pranks, fails} often get misclassified. Similarly, videos with \textit{static images} and \textit{animations} often get misclassified. 

\begin{table}
\begin{subtable}[t]{\columnwidth}
\centering
\resizebox{.9\columnwidth}{!}{
\begin{tabular}{llcccc}
\toprule
\textbf{Modality} & \textbf{Backbone} & \textbf{Top-1} & \textbf{Top-3} & \textbf{Top-5} \\
\midrule
Visual & R(2+1)D-50 & 50.6 & 72.3 & 81.4 \\
Visual & R3D-50 & 52.7 & 74.5 & 83.6 \\
Visual & R3D-101 & 52.6 & 74.1 & 83.3 \\
\midrule
Audio & VGG & 31.6 & 50.5 & 60.9 \\
Audio & CLSRIL23 & 31.2 & 50.1 & 60.6 \\
\midrule
Visual, Audio & R3D-50 + VGG & 54.9 & 74.9 & 82.4 \\
Visual, Audio & R3D-50 + CLSRIL23 & 54.9 & 75.4 & 82.9 \\
Visual, Audio & R3D-50 + VGG + CLSRIL23 & \textbf{56.5} & \textbf{76.5} & \textbf{83.8} \\
\bottomrule
\end{tabular}
}
\vspace{5pt}
\caption{\footnotesize{Concept Classification}}
\vspace{5pt}
\label{tab:topic_classification}
\end{subtable}

\begin{subtable}[t]{\columnwidth}
\centering
    \resizebox{.8\columnwidth}{!}{
    \begin{tabular}{rcccc}
     \toprule
        \textbf{Method} & \textbf{Modality} & \textbf{Top-1}& \textbf{Top-3} & \textbf{F1}\\
    \midrule
    Kosti et al.~\cite{kosti2017emotic} & visual & $35.08$ & $81.92$ & $0.19$ \\
    \midrule
     
    \multirow{2}{*}{\rotatebox{0}{Tsai et al. \cite{tsai2019multimodal}}}& audio & $27.10$ & $66.67$ & $0.21$ \\
    & audio-visual & $38.05$ & $83.90$ & $0.29$ \\
     \bottomrule
    \end{tabular}
    }
    \vspace{5pt}
    \caption{\footnotesize{Affective State Classification}}
    \label{tab:emotion_classification}
\end{subtable}
\caption{\small{\textbf{Baseline Experiments: }Baseline experiments for concept and affective state classification on \datasetname using different modalities and combinations.}}
\label{tab:table3_4}
\end{table}

\begin{table}[t]
    \centering
    \resizebox{0.7\columnwidth}{!}{
    \begin{tabular}{rcccc}
     \toprule
        \textbf{Method} &  \textbf{\#Posts} & \textbf{Top-1}& \textbf{Top-3} & \textbf{Top-5}\\
    \midrule
    Audio-Visual & - & 56.5 & 76.5 & 83.8 \\
    ProbDist & 1 & 56.9 & 77.2 & 84.1 \\
    ProbDist & 5 & 58.5 & 77.7 & 84.7  \\
    ProbDist & 20 & \textbf{59.3} & \textbf{78.9} & \textbf{85.7} \\
    ProbMax & 20 & 58.8 & 78.1 & 85.2 \\
     \bottomrule
    \end{tabular}
    }
    \caption{\small{\textbf{Creator Profiling:} Concept classification with semantically inferred creator profile with audio-visual representations.}}
    \label{tab:creator_modelling}
\end{table}
\begin{table}[t]
    \centering
    \resizebox{.8\columnwidth}{!}{
    \begin{tabular}{lccc||ccc}
    \toprule
     \textbf{Target} & \cellcolor{green!25}\textbf{Top-1} & \cellcolor{green!25}\textbf{Top-3} & \cellcolor{green!25}\textbf{Top-5} &\cellcolor{blue!25}\textbf{ Top-1} & \cellcolor{blue!25}\textbf{Top-3} & \cellcolor{blue!25}\textbf{Top-5} \\
     \midrule
    Hindi & 40.1 & 61.5 & 70.5 & 61.2 & 79.1 & 85.5\\
    Telugu & 48.1 & 72.1 & 81.6  & 54.9 & 78.2 & 86.1\\
    Tamil & 45.8 & 66.6 & 78.1 & 51.0 & 73.8 & 82.3 \\
    Kannada & 48.5 & 72.7 & 79.6 & 56.8 & 78.4 & 84.4\\
    Punjabi & 39.9 & 62.2 & 72.9 & 45.7 & 69.5 & 79.4 \\
    \bottomrule
    \end{tabular}
    }
    \caption{\small{\textbf{Cross-lingual Experiments: } We train the audio-visual concept classification model on all the languages apart from the target language and evaluate on target language~(green column) columns; all languages are used for training~(blue column). }}
    \label{tab:cross_lingual}
\end{table}
\vspace{-10pt}
\section{Social Media Content Analysis}
\label{sec:tasks}
\noindent\textbf{Creator User Profile Modeling:} \label{subsec:baseline-creator-modelling:} We leverage affinity of creators towards concepts for improving semantic understanding in Table~\ref{tab:creator_modelling}. For every creator, we mine recent videos uploaded by them and use our audio-visual semantic model for predicting the concept probabilities for these posts. We average the predicted probability distributions and use them for representing the creator (ProbDist). Creator representation is then combined with audio-visual features via late fusion for training the model. We observe gains of $5\%$ over the audio-visual baseline by incorporating creator profile as prior for semantic understanding. We vary the number of recent posts and observe gains by increasing the number of posts~(Row $2,3,4$), showing that longer creation history is helpful in modeling the creators. We also experiment with maximum prediction (ProbMax) for each post instead of probability distribution~(Row $5$). This simple yet effective baseline motivates further investigation for modeling creator user profiles using only semantics.

\noindent\textbf{Cross-Lingual Analysis:} We also explore \datasetname for cross-lingual analysis over $5$ popular languages in Table~\ref{tab:cross_lingual}. For each target language, we remove it from the training set and train an audio-visual model using other languages. We evaluate this model on the target language to obtain zero-shot results. We present the top-1, top-3, and top-5 accuracy for concept classification with this experiment in \textit{green} columns. In \textit{blue} columns, we use all $5$ languages for training and testing. We can see that the performance gap between \textit{green} and \textit{blue} columns is significant, indicating that \datasetname can be useful for advancing the state-of-the-art in cross-lingual video understanding tasks.

\noindent\textbf{Temporal Analysis:}
We explore another interesting aspect of \datasetname - temporally evolving content. We notice a strong link to real-world events~(Figure~\ref{fig:trends}). We extract top-performing $50$K posts based on views from $10$ weeks (29th August - 7th November 2021) and analyze the predictions for these posts using our models. We observe an increasing content related to \textit{sports} concept because of an upcoming major sports league. Similarly, we see some peaks in \textit{celebrations} concept because of the recent festive season. 

\section{Ethics, Data and User Privacy}
\label{sec:data_privacy}

\noindent\textbf{Respecting User Privacy:} \datasetname videos are publicly available on \textit{Moj}. Informed consent of the users has been taken by the platform for public usage of these videos. The user identifiers and exact publication date have been masked to protect privacy. 
\\
\noindent\textbf{Respecting Intellectual Property:} Creators have the complete freedom to take down their content. Our dataset provides direct URL links to access the videos, while the platform holds the rights to these videos. This would allow
the users to delete the videos on the platform, thus deactivating the links. Our data collection and dissemination efforts abide by platform guidelines.

\noindent\textbf{Opt-out form:} Users may choose to have their video removed from the dataset upon request through
an opt-out form is available on the dataset homepage.

\noindent\textbf{Handling Misuse:} Adequate caution was taken to not store any user information, videos (raw or processed), or meta-data on permanent storage outside the computing infrastructure of the social media platform. We aim to disseminate the data upon request and log all access to the dataset, which will only be available for research purposes. 

\noindent\textbf{License:} We release 3MASSIV for research purposes.

\noindent\textbf{Annotator Compensation:}
We ensured that all annotators were fairly compensated on an hourly basis and they were apprised of potential social media fatigue \cite{Zhang2021TheDA} resulting from long exposure to social media content. 

\section{Conclusion}
We presented~\datasetname, a multilingual, multimodal and multi-aspect, human-annotated dataset of social media short videos extracted from a social media platform. \datasetname~comprises of $50$K labeled short videos and $100$K unlabeled short videos from a popular social media platform in $11$ different languages. \datasetname~is useful to further semantic understanding of social media content which embodies unique characteristics and nuances. We presented an in-depth analysis and showed the challenges and uniqueness of the dataset using baseline comparisons. We also present some applications of \datasetname~for various user-modeling tasks and cross-lingual tasks. 

\section{Acknowledgements}
Mittal, Mathur, Bera and Manocha were supported, in part by ARO Grants W911NF1910069 and W911NF2110026.

{\small
\bibliographystyle{ieee_fullname}
\bibliography{egbib}
}
\clearpage
\newpage
\newpage
\begin{appendix}
\section{Dataset Details}
\label{sec:appendix-dataset-details}
\subsection{Annotation Label Descriptions}
\label{subsec:annotation-label-description}
We present descriptions of all the annotation labels of the $4$ taxonomies in Table~\ref{tab:label-taxonomy-description-big-table}. We shared these descriptions and sample videos to the annotators for reference. 
\subsection{Dataset}
\label{subsec:dataset-cleaning}

\noindent\textbf{Creator Privacy: }We have taken due measures for safeguarding the privacy of people present in our dataset. We will only be releasing the public URLs for the videos which would allow the creators to delete their videos if they do not wish to be included in the dataset at any point of time. We have masked the user identifiers and the date of upload of these videos also to protect privacy. Additionally, these videos are already publicly available on the platform and we did not augment the videos with any information which is not in the public domain. 

\noindent\textbf{User Consent: } To further our efforts towards user privacy, we have created an opt-out form where users can reach out to the team and have their videos removed from the dataset. The dataset would be available for only research purposes and commercial usage of the dataset will be strictly prohibited. 

\noindent\textbf{Geographical and Linguistic Diversity: } Our dataset contains videos from $11$ Indic languages which alludes to the high linguistic diversity. However, we understand that the dataset is dominated towards certain population and does not represent the global demographics. Similar observations have been made by Piergiovanni et al.\footnote{https://arxiv.org/pdf/2007.05515.pdf} for large scale video datasets like HVU, HACS and Kinetics where majority of the videos belong to North America with less representations for Asia, Africa, Europe and Latin America. We encourage follow-up works to be cognizant of the demographics of our dataset. As part of future work, we intend to substantially increase our dataset and increase the coverage to more languages and countries.

\begin{figure}[h]
    \centering
    \includegraphics[width =\columnwidth]{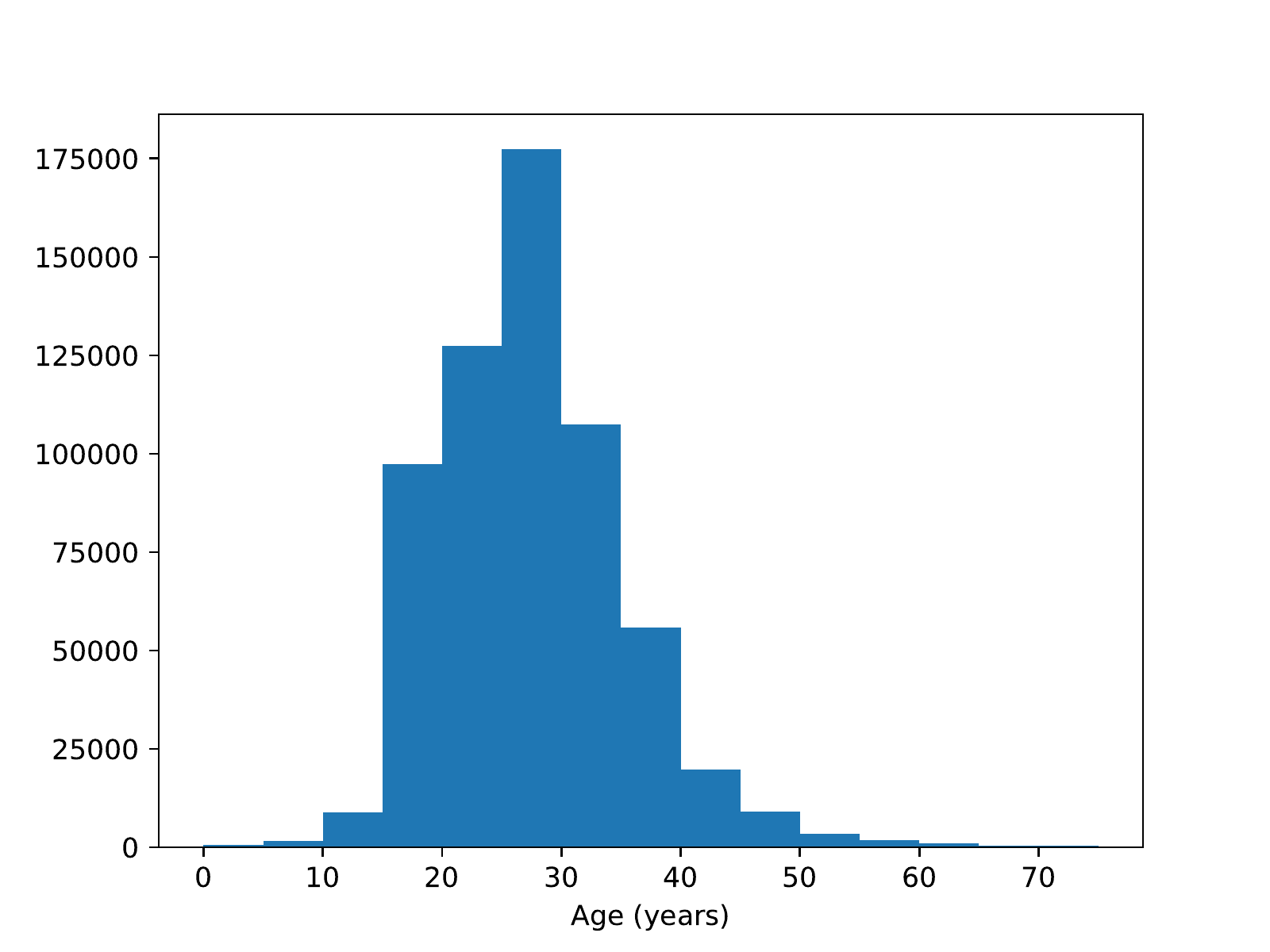}
    \caption{Distribution of age (years) of using the faces appearing in \datasetname. Average age of \datasetname computes to 27 years.}
    \label{fig:age}
\end{figure}

\noindent\textbf{Age and Gender Bias: } We extract faces from our videos using an off-the-shelf facial analysis library\footnote{https://github.com/deepinsight/insightface}. We note a healthy male-to-female ratio of 3:2 highlighting gender diversity of our dataset. The average age computes to 27 years (7 years as standard deviation) and ranges from infants to over 80 years demonstrating the age diversity of \datasetname. In Figure~\ref{fig:age}, we plot the age distribution of our dataset. We understand that this analysis is not entirely accurate due to the mis-classification of the facial analysis models but it helps to understand the overall age and gender representation of the dataset.

\noindent\textbf{Offensive Videos: }Since the videos uploaded on the platform are user-generated, there are fair chances of uploading offensive videos also. The content moderation pipeline of the platform already filters out majority of the offensive videos. To further ensure a clean dataset, our expert annotation team flagged all the videos containing offensive content like hate-speech, pornography, explicit or suggested nudity, violence (guns/knives), gore, not-suitable-for-work (NSFW), abuses etc. We removed all of these videos from our dataset. 

\noindent\textbf{De-duplication: }On social media, popular and viral videos are usually re-uploaded with minor or no changes resulting in lot of similar videos.  We used de-duplication strategies to remove these visually similar duplicate videos from the dataset. The de-duplication was performed by extracting the visual features from the videos using 3D ResNet~\cite{hara2018can}) and removing the video samples having more than $0.8$ cosine similarity scores. We employed this de-duplication to promote more diversity in the dataset. After this filtering, we ended up with a corpus of $50$K videos that were then annotated by domain experts.

\subsection{Creator Profiling}
For modelling the creators of the videos, we select recent posts uploaded by them on the platform. These posts were selected such that they do not have an overlap with the labelled $50k$ videos. We use the global average of the creator representations for the creators with insufficient historical posts. 

\subsection{In-depth Inter-Annotator Agreement}
\label{subsec:in-depth-annotator-agreement}

We summarize the per-label and per-taxonomy inter-annotator agreement values in Table~\ref{tab:annotator-agreement-big-table}. In general, inter-annotator agreement for all labels in theme taxonomy is high. Compared to other aspects, affective states show the most contrast in inter-rater confidence scores. This phenomenon has been observed in past works as well that have shown that infrequent emotions can have relatively high inter-rater correlation and frequent emotions can have have relatively low inter-rater correlation~\cite{demszky2020goemotions}. Similarly, for the audio type and video type, it is evident from Table~\ref{tab:annotator-agreement-big-table} that labels that are infrequent have lower agreement values. 

\subsection{More Data Analysis}
\label{subsec:more-data-analysis}
\begin{figure}[t]
\centering
    \begin{subfigure}[h]{\columnwidth}
        \includegraphics[width=\textwidth]{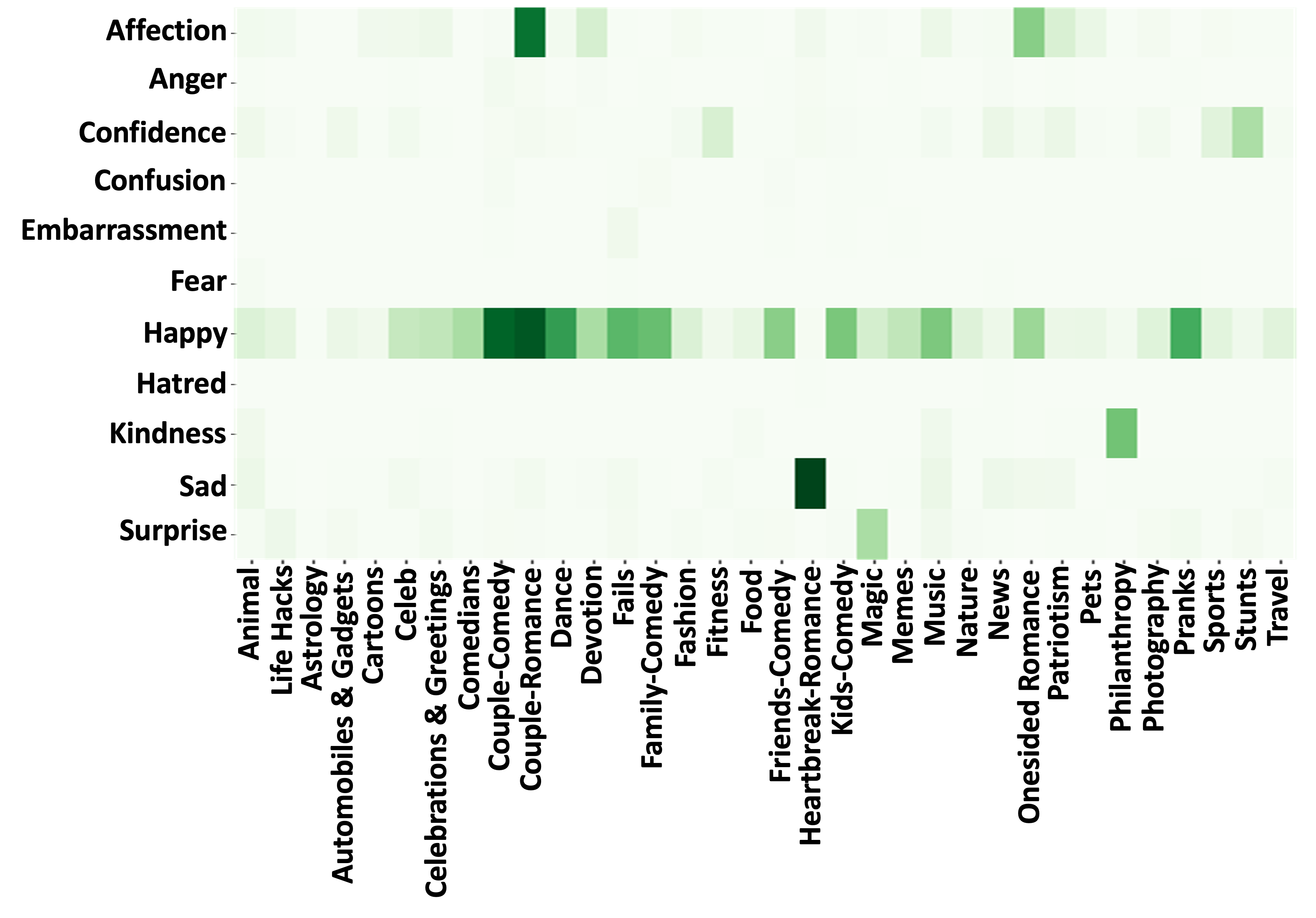}
        \caption{\textbf{Correlation Heatmap: }Correlation plot between affective state and concepts. We note high correlation between "Heartbreak-Romance" and "Sad". "Magic" correlates with "Surprise". and "Couple-Romance" with "Affection".}
        \label{fig:emotion-theme}
    \end{subfigure}
   \begin{subfigure}[h]{\columnwidth}
    \includegraphics[width=\textwidth]{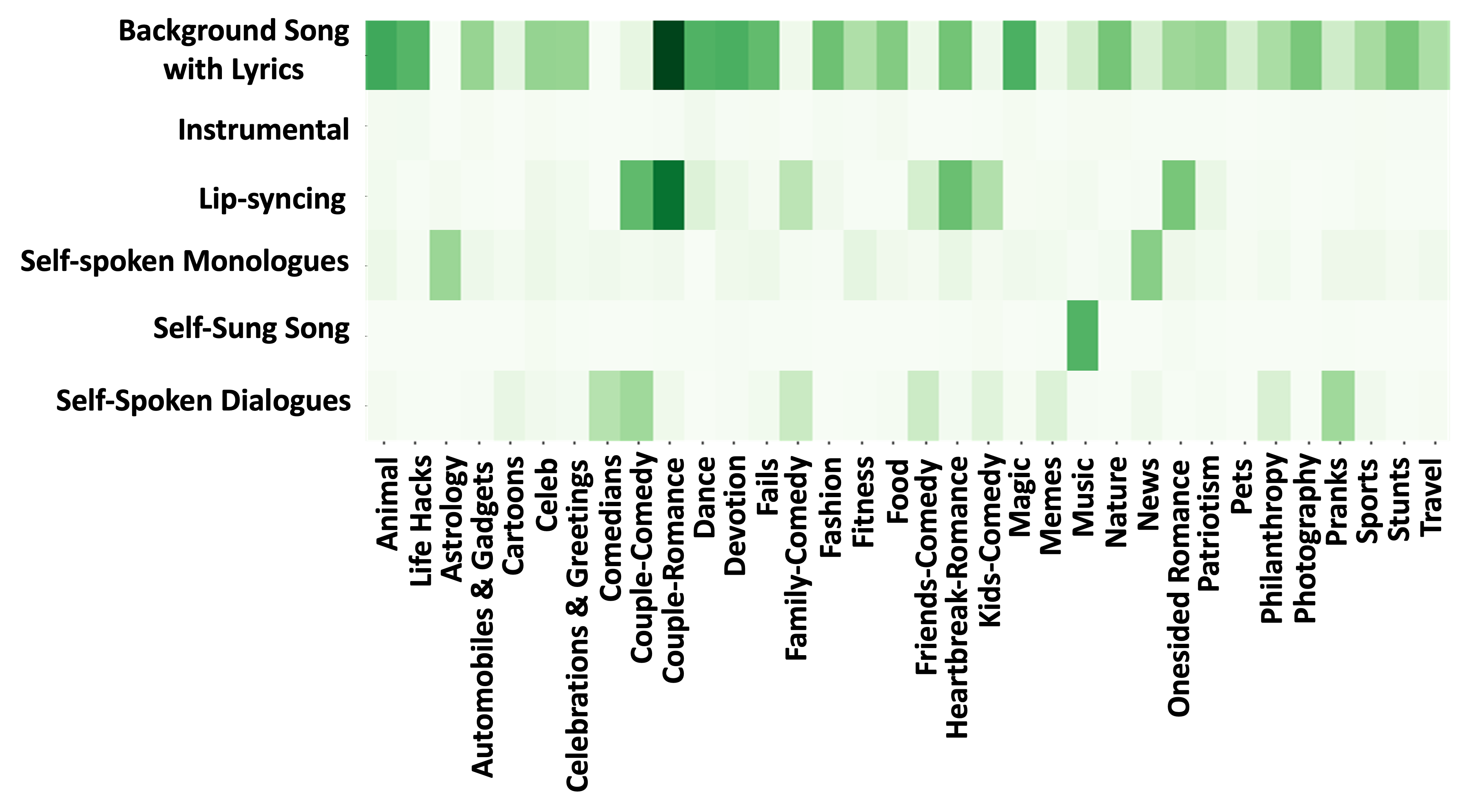}
    \caption{Correlation between \textbf{audio-type} and \textbf{concept}}
    \label{fig:audio-theme}
  \end{subfigure}
 \begin{subfigure}[h]{\columnwidth}
    \includegraphics[width=\textwidth]{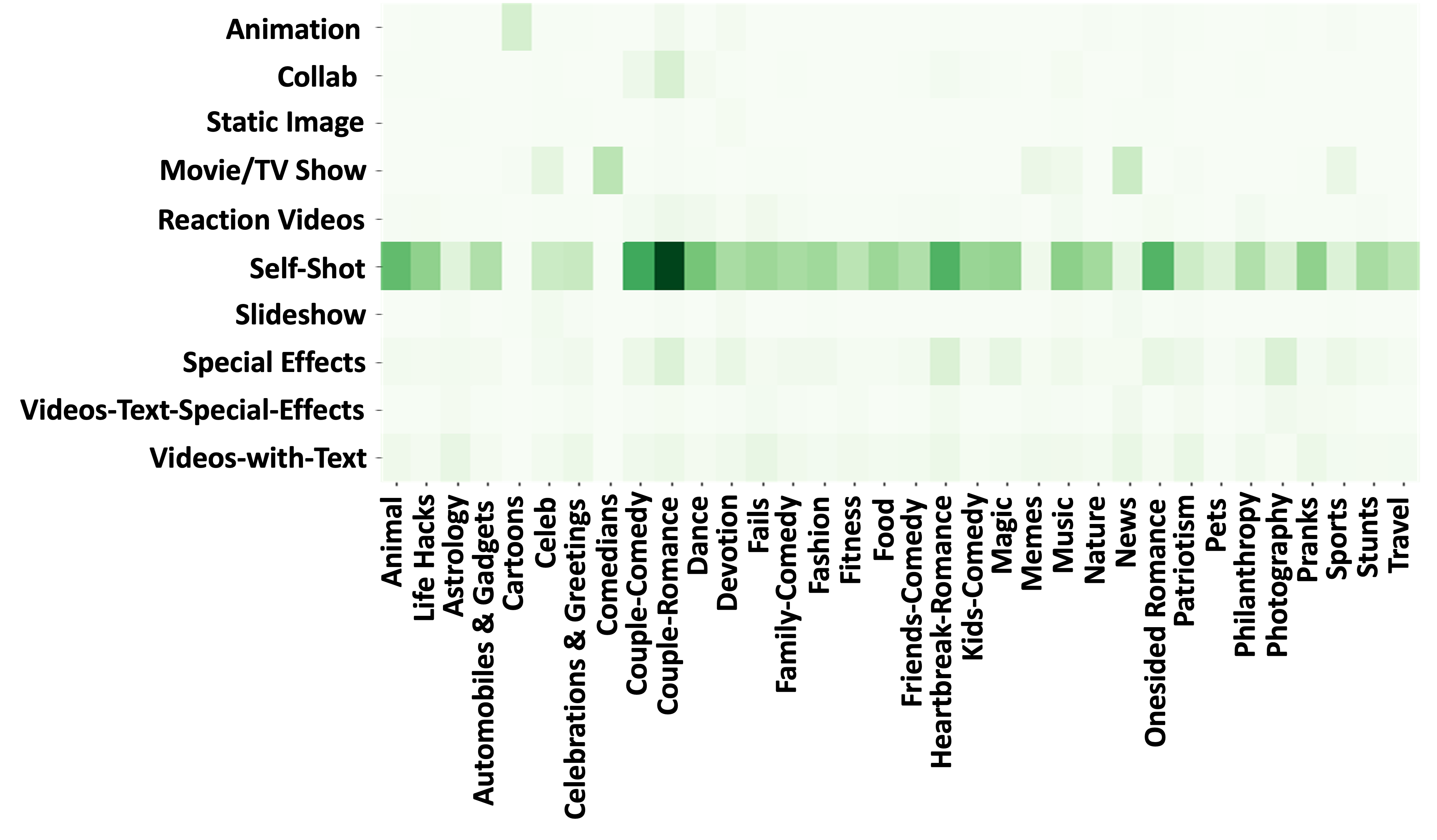}
    \caption{Correlation between \textbf{video-type} and \textbf{concept}}
    \label{fig:video-theme}
  \end{subfigure}
\caption{Correlation heatmap between different taxonomies of \datasetname.}
  \label{fig:correlation-heatmap-additional}
\end{figure}

\begin{figure}[t]
    \centering
    \includegraphics[width =\columnwidth]{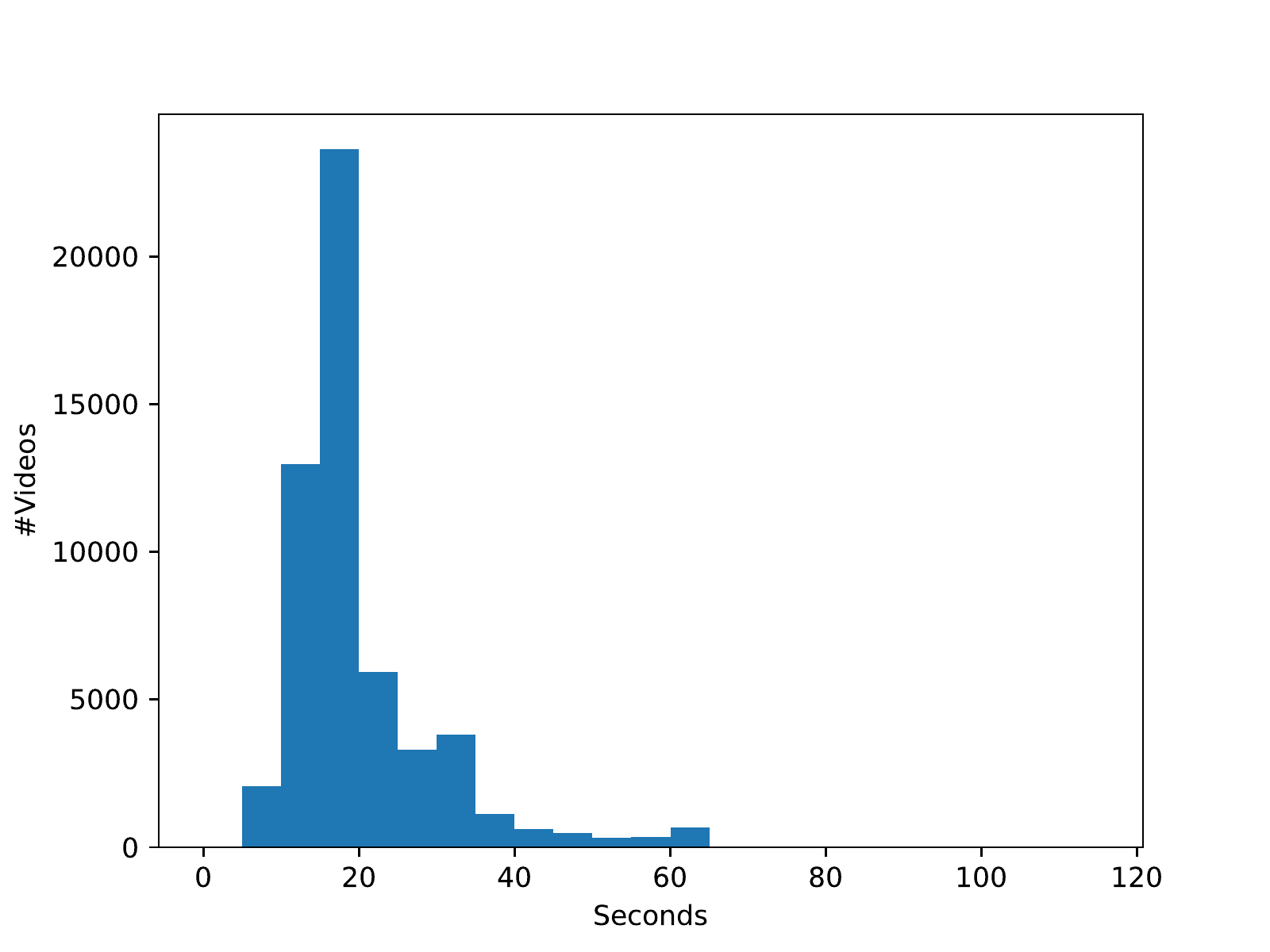}
    \caption{Distribution of video durations of \datasetname}
    \label{fig:video_dur}
\end{figure}

Figure~\ref{fig:audio-theme} and Figure~\ref{fig:video-theme} presents the correlations between audio type and theme labels and video type and theme labels respectively. We observe that \textit{comedy} and \textit{romance} show a high co-occurrence with \textit{lip-syncing} videos, while a large majority of \textit{music videos} have \textit{self-sung songs}. Similarly, it can be observed that \textit{news} category is mostly composed of \textit{self-spoken monologues} as the creators tend to use a narrative style for such content. \textit{Pranks} and \textit{comic} scenes are often presented as \textit{self-spoken dialogues} as they generally involve at least two or more people. Most videos in our dataset are \textit{self-shot}. Videos featuring \textit{comedian}s and \textit{trending news} are generally directly sourced from \textit{movie/TV show clips} which finds evidence in Figure~\ref{fig:video-theme}. In Figure~\ref{fig:video_dur}, we present the distribution of the length of videos in seconds. Our videos range from 5 seconds to 116 seconds with an average duration of 20 seconds.

\subsection{Hashtag Analysis}
\label{subsec:hashtag-analysis-appendix}
For the $50$K annotated posts, we analyzed the co-relation between the most common hashtags and \concept taxonomy after removing common/noisy tags like \textit{"trending", "hot", "viral"} using TF-IDF. We note that 40\% of our videos did not have hashtags reiterating the role of a curated taxonomy for semantic understanding. Few tags which aligned with our taxonomy eg. "comedy" and "fun" were sparse and moreover, they were not specific enough to distinguish between the challenging yet popular categories like Kids-Humour vs Family-Humour. For affective states, audio and media type, we did not find relevant hashtags. Since social media platforms spread multiple geographies, we note that lot of hashtags are also code-mixed in nature. Our expertly curated taxonomy and annotations helped us in mitigating these problems with hashtags.



\begin{figure*}[t]
     \centering
     \begin{subfigure}[b]{0.48\textwidth}
         \centering
         \includegraphics[width=\columnwidth]{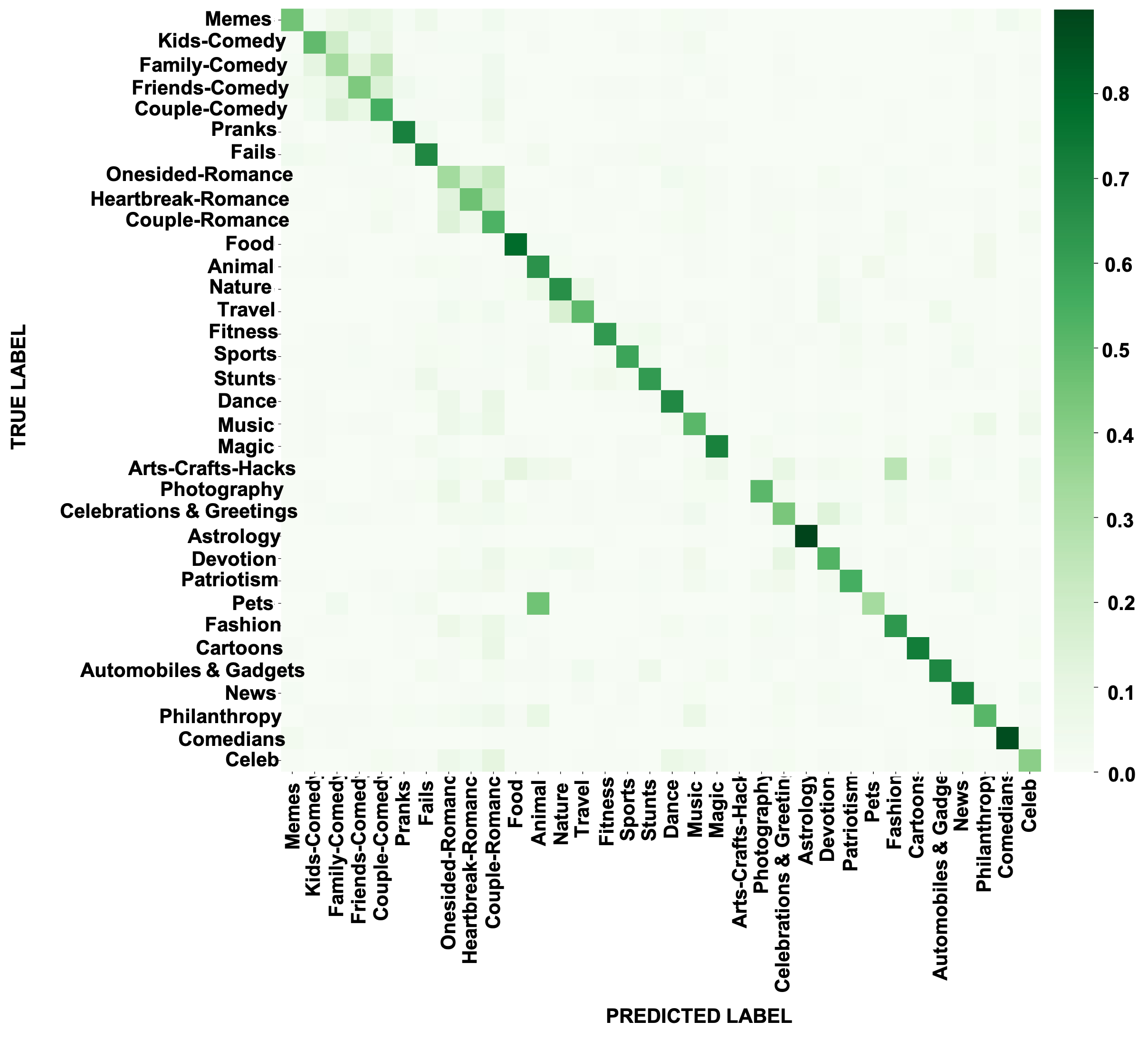}
         \caption{\textbf{Confusion Matrix for Audio-Visual Concept Classification: }This is the confusion matrix for the audio-visual concept classification experiments as discussed in Section~\ref{subsec:baseline-concept-classification}}
         \label{fig:confusion_matrix}
     \end{subfigure}
     \hfill
     \begin{subfigure}[b]{0.48\textwidth}
         \centering
         \includegraphics[width=\columnwidth]{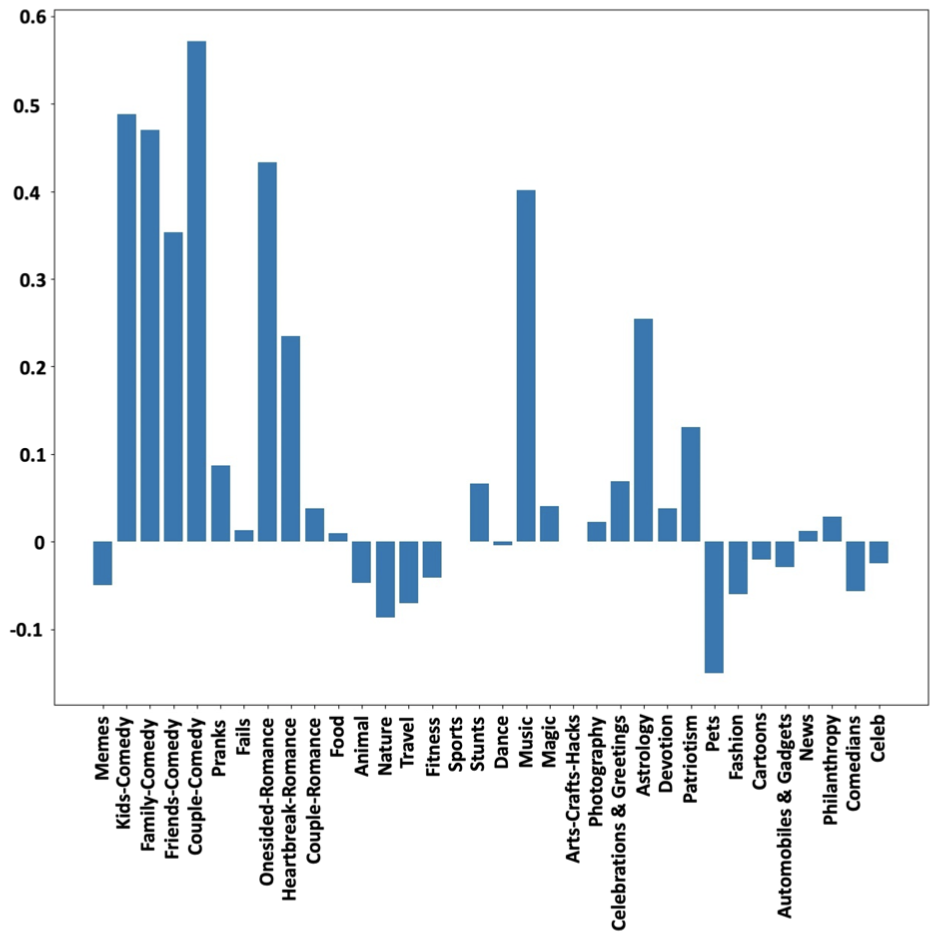}
         \caption{\textbf{Multimodal Analysis for Concept Classification: }We analyze the concept labels for which audio and the visual modalities help or degrade the accuracy of the concept classification.}
        \label{fig:audio_visual_diff}
     \end{subfigure}
     \caption{\textbf{Further Analysis Audio-Visual Concept Classification:} We present an in-depth analysis regarding the audio-visual concept classification as discussed in Section~\ref{subsec:baseline-concept-classification}.}
     \label{fig:error_analysis}
\end{figure*}

\section{Baseline Experiments: Training Details and Hyperparameters}
\label{sec:baseline-experiments-additional}
\subsection{Concept Classification}
\label{subsec:theme-baseline-additional}

\subsubsection{Feature Extraction, Backbone Architecure and Hyperparameters}
\label{subsec:theme-baseline-additional-features}
\noindent\textbf{Visual Representations: }We use ResNext~\cite{Xie2016} pretrained on Kinetics dataset as the backbone for extracting spatio-temporal features. We sample videos at $25$ frame per second and save the frames with $240$ as the size of the shortest side while maintaining the aspect ratio. We use $16$ centrally cropped frames of dimension $112$ x $112$ x $3$ as clips for extracting $2048$-dimensional ResNext features from the last convolution layer. The features across clips of the videos are averaged to generate a $2048$-dimensional representation for the video. 

\noindent\textbf{Audio Representations:} We extract the audio channel as mono-channel from our videos using ffmpeg\footnote{https://www.ffmpeg.org/}. We sample the audios at $16$kHz and use VGG~\cite{gemmeke2017audio} and CLSRIL-23~\cite{gupta2021clsril} models for audio feature extraction. In case of CLSRILS-23, the audio features are averaged across the clip to get a $512$-dimensional vector. The features extracted from VGG are $128$-dimensional vector as we tap them out before the classification layer. 

\noindent\textbf{Creator Representations:} We use our trained audio-visual model to generate the $34$ dimensional probability distribution across concepts for the videos posted by the creators. For each creators, we average the probability distribution to represent each creator by a $34$ dimensional vector.

\noindent\textbf{Architecture:} For each modality, we pass them through $2$ fully connected layers ($512 \rightarrow 512$ for creator and audio and $1024 \rightarrow 512$ for video) followed by softmax function for normalization. We concatenate these normalized outputs and pass them through a $2$ layered late fusion network ($1536 \rightarrow 512 \rightarrow num\_classes$). We use batch normalization and ReLU activation function for the linear layers. 

The network is trained for $500$ epochs using cross entropy loss with $0.5$ as dropout, $256$ as batch size and $0.005$ as learning rate. We use AdamW optimizer and decay the learning rate by 0.3 every 15 epochs. We use PyTorch\footnote{https://pytorch.org/} for all our experiments and train our models on A100 GPUs. We use the validation set for hyper-parameter tuning and early stopping and report the results on the test set. 
\subsubsection{More Analysis}
\label{subsec:theme-baseline-additional-analysis}
In Figure~\ref{fig:confusion_matrix}, we plot the confusion matrix of the audio-visual model. We notice misclassifications among the concept labels like \textit{memes, kids, family, friends and couple comedy}, demonstrating the challenges in semantic understanding of such content. Similar confusion can also be observed for \textit{romance} labels also.
To analyze further, in Figure~\ref{fig:audio_visual_diff}, we study the impact on the accuracy of concept categories after using the audio modality also. We note that the majority of the categories benefit by adding audio indication rich information encoded in the audio channel also. However, some classes which are intuitively less aligned with audio and have generic background audios like \textit{gadgets, animals, fashion, nature, travel} get impacted negatively. 

\begin{figure*}[t]
    \centering
    \includegraphics[width =0.7\textwidth]{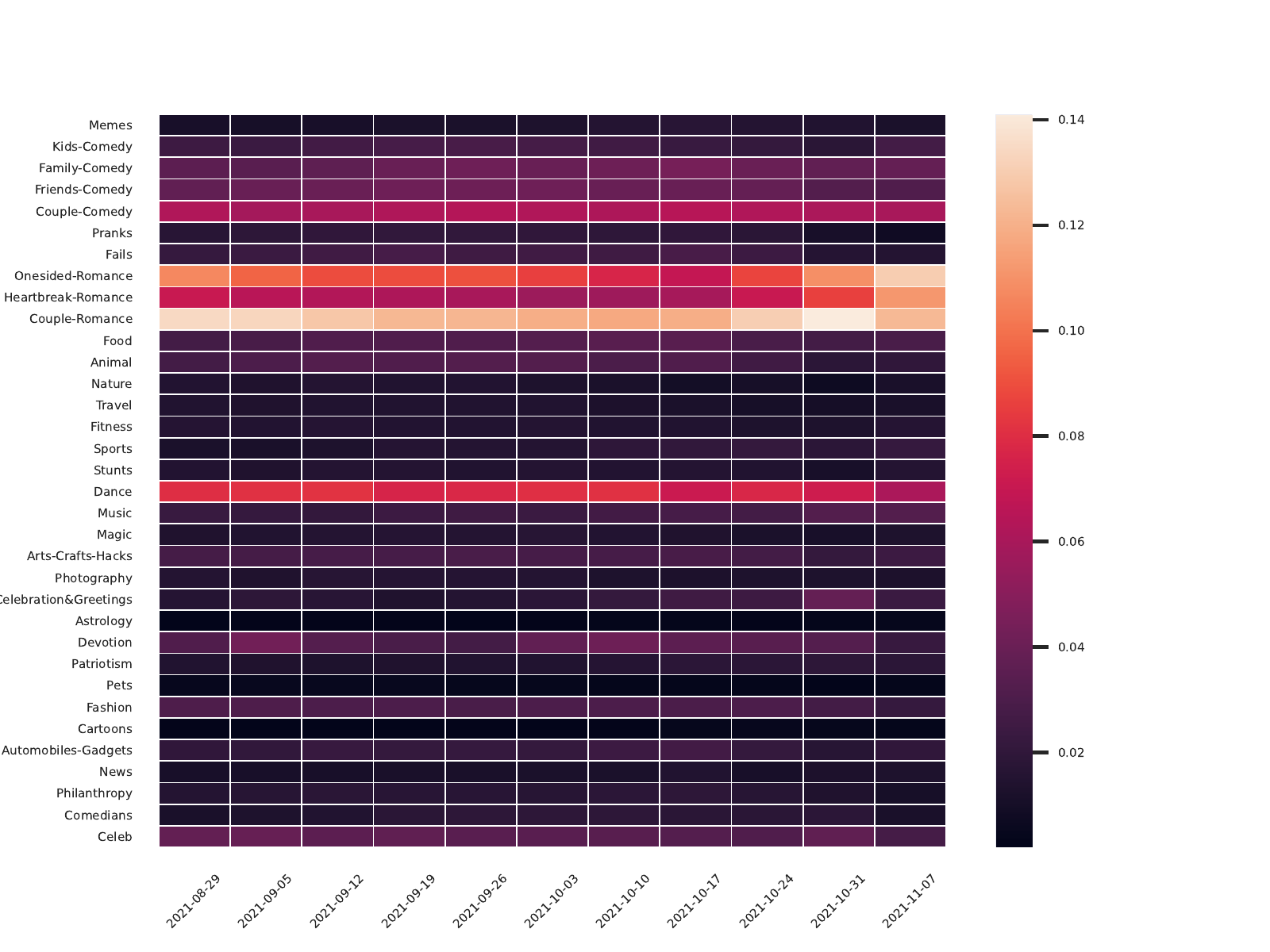}
    \caption{\textbf{Temporal Trends:} We plot the predictions of our audio-visual model on top performing post for 11 weeks (Aug, 2021 - Nov, 2021) to analyze the correlation between content uploaded and real-life events. \textit{Sports} and \textit{Festivals} show positive correlation with a sports league and festive season.}
    \label{fig:trends}
\end{figure*}

\subsection{Temporal Trends}
In Figure~\ref{fig:trends}, we collect top performing posts (~$50$k) from each week (Aug, 2021 - Nov, 2021) on the basis of number of views. We use our trained audio-visual model to predict the probability distribution for these weekly posts to understand the correlation between real-life events and content uploaded on platform. We observe higher number of posts related to sports due to an upcoming major sports league.  Similarly, we see some increase in posts related to festivals because of the recent festive season. These trends show how the temporal distribution of videos uploaded on social media platforms is dynamic and an interesting research direction.

\subsection{Broader Impact, Limitations and Potential Negative Impact}
Majority of the existing datasets source videos from social media for video understanding, but are focused towards specific tasks like action recognition, object detection/segmentation etc. Our dataset \datasetname provides a unique opportunity for modeling the dynamics behind creations and consumption of these social media videos on short video platform which is a unique, novel and popular video format. \datasetname can be a stepping stone towards improving our understanding of human behaviour and preferences on video-first social media platforms. Our annotation of affective states can be instrumental in detecting signs of stress, cyber-bullying, cruelty on social media video platforms and help in providing a safe user experience. We annotate our dataset for $11$ Indic languages is an attempt to develop an inclusive dataset, reducing the language bias and provide more representation to an under-represented population. However, this also forms the limitation of our dataset for truly global understanding, which we intend to tackle as follow-up work. Our dataset shows healthy distribution across age and gender, but also shows the natural imbalance in the age distribution owing to the adoption pattern of social media platforms across different age groups. Our dataset currently captures social media trends during a 9 month duration, which may not capture the entire range of existing trends occurring on social media. Since our dataset contains a variety of popular videos, often created by popular personalities, it could enable malicious parties to track and monitor people. However, we took substantial measures to preserve the identity of people by masking the dates and user identifiers for these videos. Additionally, we only use videos which are publicly available to prevent breach of user privacy and trust.

\begin{table*}[t]
    \centering
    \resizebox{0.75\textwidth}{!}{
\begin{tabular}{clll}
\toprule
\multicolumn{1}{l}{\textbf{Taxonomy}}                           & \textbf{Labels}                            & \textbf{Inter-Annotator Agreement} & \textbf{Avg. Inter-Annotator Agreement} \\ \midrule
                                                                  & Music                                      & 0.63                               &                                         \\
                                                                  & Animal                                     & 0.92                               &                                         \\
                                                                  & Dance                                      & 0.84                               &                                         \\
                                                                  & Devotion                                   & 0.82                               &                                         \\
                                                                  & Magic                                      & 0.93                               &                                         \\
                                                                  & News                                       & 0.91                               &                                         \\
                                                                  & Celeb                                      & 0.81                               &                                         \\
                                                                  & Fails                                      & 0.81                               &                                         \\
                                                                  & Memes                                      & 0.85                               &                                         \\
                                                                  & Life Hacks                                 & 0.86                               &                                         \\
                                                                  & Fashion                                    & 0.81                               &                                         \\
                                                                  & Food                                       & 0.9                                &                                         \\
                                                                  & Nature                                     & 0.83                               &                                         \\
                                                                  & Philanthropy                               & 0.78                               &                                         \\
                                                                  & Patriotism                                 & 0.89                               &                                         \\
                                                                  & Stunts                                     & 0.84                               &                                         \\
                                                                  & Festival                                   & 0.79                               &                                         \\
                                                                  & Sports                                     & 0.83                               &                                         \\
                                                                  & Photography                                & 0.9                                &                                         \\
                                                                  & Automobiles \& Gadgets                     & 0.85                               &                                         \\
                                                                  & Fitness                                    & 0.87                               &                                         \\
                                                                  & Travel                                     & 0.73                               &                                         \\
                                                                  & Astrology                                  & 0.96                               &                                         \\
                                                                  & Cartoons                                   & 0.85                               &                                         \\
                                                                  & Pets                                       & 0.92                               &                                         \\
                                                                  & Pranks                                     & 0.83                               &                                         \\
                                                                  & Comedians                                  & 0.68                               &                                         \\
                                                                  & \cellcolor[HTML]{FFFFFF}Couple Romance     & 0.65                               &                                         \\
                                                                  & \cellcolor[HTML]{FFFFFF}Heartbreak Romance & 0.62                               &                                         \\
                                                                  & \cellcolor[HTML]{FFFFFF}Onesided Romance   & 0.68                               &                                         \\
                                                                  & \cellcolor[HTML]{FFFFFF}Kids Comedy        & 0.8                                &                                         \\
                                                                  & \cellcolor[HTML]{FFFFFF}Couple Comedy      & 0.82                               &                                         \\
                                                                  & \cellcolor[HTML]{FFFFFF}Friends Comedy     & 0.91                               &                                         \\
\multicolumn{1}{l}{\multirow{-34}{*}{\rotatebox{90}{\textbf{Theme}}}}                                 & \cellcolor[HTML]{FFFFFF}Family Comedy      & 0.89                               & \multirow{-34}{*}{0.77}                 \\ \midrule
\multicolumn{1}{l}{}                                            & Happy                                      & 0.35                               &                                         \\
\multicolumn{1}{l}{}                                            & Affection                                  & 0.33                               &                                         \\
\multicolumn{1}{l}{}                                            & Sad                                        & 0.61                               &                                         \\
\multicolumn{1}{l}{}                                            & Confidence                                 & 0.42                               &                                         \\
\multicolumn{1}{l}{}                                            & Surprise                                   & 0.18                               &                                         \\
\multicolumn{1}{l}{}                                            & Kindness                                   & 0.72                               &                                         \\
\multicolumn{1}{l}{}                                            & Anger                                      & 0.17                               &                                         \\
\multicolumn{1}{l}{}                                            & Confusion                                  & 0.07                               &                                         \\
\multicolumn{1}{l}{}                                            & Embarrassment                              & 0.02                               &                                         \\
\multicolumn{1}{l}{}                                            & Fear                                       & 0.14                               &                                         \\
\multicolumn{1}{l}{\multirow{-11}{*}{\rotatebox{90}{\textbf{Affective State}}}} & Hatred                                     & 0.08                               & \multirow{-11}{*}{0.40}                 \\ \midrule
\multicolumn{1}{l}{}                                            & Background song with lyrics                & 0.63                               &                                         \\
\multicolumn{1}{l}{}                                            & Lip-syncing                                & 0.51                               &                                         \\
\multicolumn{1}{l}{}                                            & Self-spoken dialogues                      & 0.65                               &                                         \\
\multicolumn{1}{l}{}                                            & Self-spoken monologues                     & 0.61                               &                                         \\
\multicolumn{1}{l}{}                                            & No sound                                   & 0.38                               &                                         \\
\multicolumn{1}{l}{}                                            & Self Sung Songs                            & 0.84                               &                                         \\
\multicolumn{1}{l}{\multirow{-7}{*}{\rotatebox{90}{\textbf{Audio Type}}}}       & Instrumental Music Recording               & 0.31                               & \multirow{-7}{*}{0.59}                  \\ \midrule
\multicolumn{1}{l}{}                                            & Self-shot video                            & 0.71                               &                                         \\
\multicolumn{1}{l}{}                                            & Video with Special Effects                 & 0.52                               &                                         \\
\multicolumn{1}{l}{}                                            & Video with Text                            & 0.52                               &                                         \\
\multicolumn{1}{l}{}                                            & Splitscreen                                & 0.73                               &                                         \\
\multicolumn{1}{l}{}                                            & Movie / TV Show Clips                      & 0.67                               &                                         \\
\multicolumn{1}{l}{}                                            & Animation \& Digital Art                   & 0.72                               &                                         \\
\multicolumn{1}{l}{}                                            & Slideshow                                  & 0.44                               &                                         \\
\multicolumn{1}{l}{\multirow{-8}{*}{\rotatebox{90}{\textbf{Video Type}}}}       & Static Image                               & 0.41                               & \multirow{-8}{*}{0.62}                  \\ \bottomrule
\end{tabular}
}
\caption{\textbf{Inter Annotator Agreement:} We summarize the per-label and per-category annotator agreement for \datasetname.}
\label{tab:annotator-agreement-big-table}
\end{table*}
\clearpage
\begin{table*}[t]
    \centering
    \resizebox{0.9\textwidth}{!}{
\begin{tabular}{cll}
\toprule
\textbf{Taxonomy}                          & \textbf{Labels}                            & \textbf{Description}  \\ \midrule
\multirow{34}{*}{\rotatebox{90}{Concept}} & Music & Singing, beat-boxing, playing an instrument or other musical performance \\
& Dance  & People performing solo/group dances \\  
& Devotion & Videos related to divinity, spirituality and religion \\    
& Magic  & Magicians performing tricks and illusions \\    
& News  & Videos containing news, reports or coverage of events\\    
& Celeb  & Videos of celebrities from entertainment industry\\
& Fails  & Some unplanned event creating a humorous situations\\  
& Memes  & Viral audio/video which are slightly edited to suit different contexts\\    
& Life Hacks  &  Simple tricks for making daily activities easier\\
& Fashion  & Videos showing/making aware of fashion tricks and trends\\    
& Food  & Videos focusing on preparation/consumption/review of food or beverage\\
& Nature & Videos capturing natural scenes like rivers, trees, mountains\\  
& Philanthropy  & Selfless and kind actions like helping poor\\    
& Patriotism  & Generate the feeling of love towards country\\    
& Stunts & Showcasing challenging and thrilling skills and activities \\    
& Festivals & Celebrating and sharing greetings during various festivals\\
& Sports & Clips contain any professional sports\\  
& Animal & Videos containing wild animals  \\
& Photography & Display tricks related to photography\\    
& Automobiles \& Gadgets & Video featuring automobiles or gadgets \\    
& Fitness & Videos focusing on improving mental and physical health \\ 
& Travel & Videos capturing travel destinations and journeys\\    
& Astrology & Videos containing astrology\\    
& Cartoons & Videos with animated characters \\
& Pets & Pet animals like dogs, cats etc \\  
& Pranks & Mischievous tricks that generate humorous situations\\    
& Comedians & Popular comedian performing jokes or stand up act \\    
& Couple Romance & Videos showing romance between couples\\
& Heartbreak Romance  & People expressing feelings after breakup/betrayal \\
& One sided Romance  & Videos showing that only one person is in love\\
& Kids Comedy & Humorous acts performed majorly by kids \\    
& Couple Comedy  & Funny content about couple relationship\\    
& Friends Comedy & Funny content about friends\\ 
& Family Comedy  & Humor generated around family and its members\\  
\midrule
\multirow{11}{*}{\rotatebox{90}{Affective State}} & Happy & Videos eliciting happiness in viewers\\
& Affection & Videos depicting love and fondness\\
& Sad & Videos expressing sadness, grief, pain, suffering \\
& Confidence & People showing high confidence and self-esteem\\
& Surprise & Videos having an element of surprise\\
& Kindness & Video having an element of kindness\\
& Anger & Videos with people expressing anger and annoyance\\
& Confusion & Videos showing confusion among people\\
& Embarrassment & Videos depicting a feeling of embarrassment\\
& Fear & Videos having an element of Fear\\
& Hatred & Videos showing hatred and disapproval\\
\midrule
\multirow{7}{*}{\rotatebox{90}{Audio Type}}  & Background song with lyrics &   Background song having both lyrics and instrumentals\\
& Lip-syncing &  People lip-syncing to a pre-recorded song or dialogue\\
& Self-spoken dialogues & Two or more people conversing with each other \\
& Self-spoken monologues &  Single person speaking\\
& No sound &  Feeble noise or complete silence in the video\\
& Self-sung songs &  Audio contains song sung by the individual themselves \\
& Instrumental & Instrumental music in the background. No lyrics. \\
\midrule
\multirow{8}{*}{\rotatebox{90}{Video Type}} & Self-shot video & Original videos created by users \\
& Video with Special Effects & Videos with special artifacts like masks, blur effect etc. \\
& Video with Text &  Text superimposed on the video\\
& Splitscreen  & Two or more screens placed side by side\\
& Movie / TV Show Clips  & One or multiple shots compiled together from an existing movie or TV-show\\
& Animation \& Digital Art & Visuals generated using digital technology \\
& Slideshow  &  Sequence of images/video-snippets combined together as a video\\
& Static Image  & Static image throughout the video\\ \bottomrule
\end{tabular}
}
\caption{\textbf{Taxonomy Descriptions:} We summarize the instructions shared with the annotation team for annotation \datasetname videos.}
\label{tab:label-taxonomy-description-big-table}
\end{table*}

\end{appendix}
\end{document}